\RequirePackage{luatex85}
\RequirePackage[OT1]{fontenc}
\documentclass[journal, twoside]{IEEEtran}

\newif\ifexternalize
\externalizefalse

\usepackage{array,amsthm}
\usepackage{graphicx}
\usepackage{pgfplots}
\usepgfplotslibrary{colorbrewer}
\usepgfplotslibrary{groupplots}
\usepackage{tikz,tikz-3dplot}
\usetikzlibrary{svg.path}
\usetikzlibrary{positioning}
\usetikzlibrary{calc}
\usetikzlibrary{arrows.meta}
\usetikzlibrary{shapes.arrows, shapes.geometric}
\usetikzlibrary{backgrounds}
\usetikzlibrary{external}
\usetikzlibrary{intersections}
\ifexternalize
\fi
\usepackage{scalerel}
\usepackage{tabularx,booktabs}
\usepackage{siunitx}
\usepackage[super]{nth}
\usepackage{makecell}
\usepackage{amssymb}
\usepackage{amsmath}
\usepackage{array}
\usepackage{multirow}
\usepackage{float}
\usepackage{textcomp}

\usepackage[caption=false,font=normalsize,labelfont=sf,textfont=sf]{subfig}

\DeclareMathOperator{\vecfrommat}{vec}
\DeclareMathOperator*{\argmin}{argmin}
\DeclareMathOperator*{\argmax}{argmax}

\newcolumntype{H}{>{\setbox0=\hbox\bgroup}c<{\egroup}@{}}
\newcolumntype{Z}{>{\setbox0=\hbox\bgroup}c<{\egroup}@{\hspace*{-\tabcolsep}}}
\newcolumntype{P}[1]{>{\centering\arraybackslash}p{#1}}

\newcommand{\liegrouprigid}[1]{\mathrm{SE}(#1)}

\newcommand{\liegrouprotation}[1]{\mathrm{SO}(#1)}

\newcommand\solidrule[1][.3cm]{\rule[0.5ex]{#1}{.6pt}}
\newcommand\dashedrule{\mbox{%
    \solidrule[.5mm]\hspace{.5mm}\solidrule[.5mm]\hspace{.5mm}\solidrule[.5mm]}}

\pgfplotsset{compat=1.16}

\usepackage[hidelinks,plainpages]{hyperref}

\definecolor{oddbouncecolor}{rgb}{0,1,0}
\definecolor{evenbouncecolor}{rgb}{1,0,1}

\newcommand{\multilinecell}[2][c]{%
  \begin{tabular}[#1]{@{}l@{}}#2\end{tabular}}

\newcommand\copyrighttext{%
  \footnotesize \textcopyright 2024 IEEE. Personal use of this material is permitted. Permission from IEEE must be obtained for all other uses, in any current or future media, including reprinting/republishing this material for advertising or promotional purposes, creating new collective works, for resale or redistribution to servers or lists, or reuse of any copyrighted component of this work in other works.
  DOI: \href{https://doi.org/10.1109/TITS.2024.3397075}{10.1109/TITS.2024.3397075}}
\newcommand\copyrightnotice{%
{\tikzexternaldisable%
\begin{tikzpicture}[remember picture,overlay]
\node[anchor=south,yshift=4pt] at (current page.south) {\fbox{\parbox{\dimexpr\textwidth-\fboxsep-\fboxrule\relax}{\copyrighttext}}};
\end{tikzpicture}}%
}

\begin{document}
\IEEEoverridecommandlockouts
\def\numberposes{X}
\def\numberlandmarks{L}
\def\graphvariables{\mathbf{\Theta}}
\def\posenode{\mathbf{x}}
\def\poseindex{x}
\def\measurement{\mathbf{z}}
\def\landmarknode{\mathbf{l}}
\def\landmarkindex{l}
\def\linelandmarknode{\landmarknode^{\textnormal{l}}}
\def\pointlandmarknode{\landmarknode^{\textnormal{p}}}
\def\maplinelandmarknode{\mathbf{m}^{\textnormal{l}}}
\def\mappointlandmarknode{\mathbf{m}^{\textnormal{p}}}
\def\startnodesubscript{\textnormal{s}}
\def\endnodesubscript{\textnormal{e}}
\def\graphfactor{\mathbf{f}}
\def\graphfactorposepointlandmark{\graphfactor^{\posenode,\pointlandmarknode}}
\def\graphfactorposepose{\graphfactor^{\posenode,\posenode}}
\def\graphfactorpointlandmark{\graphfactor^{\pointlandmarknode}}
\def\graphfactorlinelandmark{\graphfactor^{\linelandmarknode}}
\def\graphfactorposelinelandmark{\graphfactor^{\posenode,\linelandmarknode}}
\def\measurementlinelandmark{\measurement^{\linelandmarknode}}
\def\measurementposepointlandmark{\measurement^{\posenode,\pointlandmarknode}}
\def\measurementpointlandmark{\measurement^{\pointlandmarknode}}
\def\measurementposepose{\measurement^{\posenode,\posenode}}
\def\errorfunction{\mathbf{e}}

\def\informationmatrix{\mathbf{P}}
\def\informationmatrixposepose{\informationmatrix^{\posenode,\posenode}}
\def\informationmatrixposepointlandmark{\informationmatrix^{\posenode,\pointlandmarknode}}
\def\informationmatrixpointlandmark{\informationmatrix^{\pointlandmarknode}}
\def\informationmatrixlinelandmark{\informationmatrix^{\linelandmarknode}}
\def\sigmapointlandmark{\sigma_{\pointlandmarknode}}
\def\sigmalinelandmark{\sigma_{\linelandmarknode}}
\def\sigmaposeposetranslation{\sigma_{\posenode,\posenode_\textnormal{T}}}
\def\sigmaposeposerotation{\sigma_{\posenode,\posenode_{\phi}}}
\def\sigmaposepointlandmarkrange{\sigma_{\posenode,\pointlandmarknode_\textnormal{R}}}
\def\sigmaposepointlandmarkbearing{\sigma_{\posenode,\pointlandmarknode_{\phi}}}

\tikzset{
  pics/carc/.style args={#1:#2:#3}{
    code={
      \draw[pic actions] (#1:#3) arc(#1:#2:#3);
    }
  }
}

\tikzset{
  mechanically/.pic={
      \foreach \d in  {0,1,...,2}
      \draw [very thick, line cap=round] (0, 0) pic{carc=45:315:0.1+0.1*\d};
      \draw [line width=0.06cm, line cap=round] (0, 0) -- (0.25, 0.25);
  }
}
\tikzset{
  polarimetricmechanically/.pic={
    \foreach \d in  {0,1,...,2}
    \draw [very thick, line cap=round] (0, 0) pic{carc=45:315:0.1+0.1*\d};
    \draw [line width=0.06cm, line cap=round] (0, 0) -- (0.25, 0.25);
    \draw [thick] (0.3, -0.25) -- ++(0, 0.2);
    \draw [thick] (0.2, -0.15) -- ++(0.2, 0);
  }
}
\tikzset{
  automotive/.pic={
      \fill (0,-0.2) circle (0.075);
      \foreach \d in  {0,1,2,3}
      \draw[very thick, line cap=round] (0,-0.2) pic{carc=45:135:.2+.1*\d};
  }
}
\tikzset{
  polarimetricautomotive/.pic={
    \fill (0,-0.2) circle (0.075);
    \foreach \d in  {0,1,2,3}
    \draw[very thick, line cap=round] (0,-0.2) pic{carc=45:135:.2+.1*\d};
    \draw [thick] (0.3, -0.25) -- ++(0, 0.2);
    \draw [thick] (0.2, -0.15) -- ++(0.2, 0);
  }
}
\tikzset{
  gpr/.pic={
    \fill (0,0.2) circle (0.075);
    \foreach \d in  {0,1,2}
    \draw[very thick, line cap=round] (0,0.2) pic{carc=225:315:.2+.1*\d};
    \draw[thick] (-0.4, -0.3) -- ++(0.8, 0);
  }
}
\tikzset{
  camera/.pic={
    \draw [thick, rounded corners=0.5mm] (-0.4, -0.25) rectangle (0.2, 0.25);
    \draw [thick] (0.2, 0.1) -- ++(0.2, 0.1) -- ++(0, -0.4) -- ++(-0.2, 0.1);
  }
}
\tikzset{
  coloredcamera/.pic={
    \draw [thick, rounded corners=0.5mm, fill=Paired-G, draw=Paired-H] (-0.4, -0.25) rectangle (0.2, 0.25);
    \draw [thick, fill=Paired-G, draw=Paired-H] (0.2, 0.1) -- ++(0.2, 0.1) -- ++(0, -0.4) -- ++(-0.2, 0.1);
  }
}
\tikzset{
  lidar/.pic={
    \begin{scope}[xshift=0.1cm]
      \fill (0,0) circle (0.1);
      \foreach \d in  {0,1,...,12}
      \draw[line width=0.2mm] (0,0) -- (30*\d:0.3);
      \foreach \d in  {0,1,...,12}
      \draw[line width=0.2mm] (0,0) -- (15+30*\d:0.2);
      \draw[thick] (0, 0) -- ++(-0.5, 0);
    \end{scope}
  }
}
\tikzset{
  gyro/.pic={
    \draw [thick] (0, 0) circle (0.3);
    \draw [rotate=30] (0,0) ellipse (0.3 and 0.1);
    \draw [rotate=120] (0,0) ellipse (0.2 and 0.1);
    \path (0, 0) -- (120:0.4) coordinate (A);
    \draw [thick] (A) -- (-60:0.4);
  }
}
\tikzset{
  accelerometer/.pic={
    \draw [thick, latex-latex] (-0.4, 0) -- (0.4, 0);
    \draw [thick, latex-latex] (0, -0.4) -- (0, 0.4);
    \draw [thick, stealth-stealth] (-0.3, -0.3) -- (0.3, 0.3);
  }
}
\tikzset{
  speedometer/.pic={
    \begin{scope}[yshift=-0.2cm]
    \draw[thick] (0, 0) pic{carc=0:180:.4};
    \foreach \d in  {0,1,...,4}
    \path (0, 0) -- (45*\d:0.4) edge [draw, thick] (45*\d:0.3);
    \fill (0,0) circle (0.05);
    \fill (40:0.05) -- (70:0.3) -- (100:.05);
  \end{scope}
  }
}
\tikzset{
  steeringangle/.pic={
    \draw[line width=0.6mm] (0, 0) circle (0.3);
    \path [fill] (175: 0.3) to [in=150, out=30] (5: 0.3) to (-5: 0.3) to
    [out=180, in=80] (-85: 0.3) to (-95: 0.3) to [out=100, in=0] (-175:0.3);
    \draw [fill, white] (0, 0) circle (0.05);
  }
}
\tikzset{
  gnss/.pic={
    \foreach \d in  {0, 1}
    \draw[very thick, line cap=round] (0,0) pic{carc=0:-90:.2+.1*\d};
    \draw [fill] (0, 0) circle (0.1);
    \path (135: 0.1) coordinate (A);
    \draw [fill] (A) circle (0.1);
    \draw [fill] (45: 0.1) -- ++(135: 0.1) -- ++(-135: 0.2) -- ++(-45: 0.1);
    \path (A) -- ++(45: 0.1) -- ++(90: 0.05) -- ++(45: 0.3) coordinate (winga);
    \path (A) -- ++(-135: 0.1) -- ++(180: 0.05) -- ++(-135: 0.3) coordinate (wingb);
    \draw [thick] (0, 0) -- ++(45: 0.1) -- ++(0: 0.05) -- ++(45: 0.3) -- (winga)
    -- ++(-135: 0.3) -- ++(-90: 0.05) -- (A);
    \draw [thick] (0, 0) -- ++(-135: 0.1) -- ++(-90: 0.05) -- ++(-135: 0.3) -- (wingb)
    -- ++(45: 0.3) -- ++(0: 0.05) -- (A);
  }
}
\tikzset{
  magnetometer/.pic={
    \draw [very thick] (0, 0) circle (0.3);
    \foreach \d in  {0,1,...,4}
    \draw (90*\d: 0.25) -- (90*\d:0.3);
    \path [fill] (45: 0.08) -- (135: 0.275) -- (-135: 0.08) -- (-45: 0.275);
    \draw [fill, white] (0, 0) circle (0.03);
  }
}
\tikzset{
  crossmodal/.pic={
    \path (-0.2, -0.25) rectangle (0.5, 0.25); %

    \draw [thick, latex-latex] (-0.2, -0.15) -- ++(0.25, 0) -- ++(0.2, 0.3) -- ++(0.25, 0);
    \draw [thick, latex-latex] (-0.2, 0.15) -- ++(0.25, 0) -- ++(0.2, -0.3) -- ++(0.25, 0);
  }
}
\tikzset{
  monomodal/.pic={
    \path (-0.2, -0.25) rectangle (0.5, 0.25); %
    \draw [thick, latex-latex] (-0.2, -0.15) -- ++(0.7, 0);
    \draw [thick, latex-latex] (-0.2, 0.15) -- ++(0.7, 0);
  }
}
\tikzset{
  dense/.pic={
    \foreach \x in  {0,1,...,2}
    \foreach \y in  {0,1,...,2}
    \draw [thick] (-0.2 + 0.2 * \x, -0.2 + 0.2 * \y) circle (0.075);
  }
}
\tikzset{
  quasidense/.pic={
    \draw [thick] (-0.2, -0.2) circle (0.075);
    \draw [thick] (0.0, -0.2) circle (0.075);
    \draw [thick] (-0.2, 0.0) circle (0.075);
    \draw [thick] (0.0, 0.0) circle (0.075);
    \draw [thick] (0.2, 0.0) circle (0.075);
    \draw [thick] (-0.2, 0.2) circle (0.075);
    \draw [thick] (0.2, 0.2) circle (0.075);
  }
}
\tikzset{
  sparse/.pic={
    \draw [thick] (-0.2, -0.2) circle (0.075);
    \draw [thick] (0.2, -0.2) circle (0.075);
    \draw [thick] (0.0, 0.0) circle (0.075);
    \draw [thick] (0.0, 0.2) circle (0.075);
  }
}

\newcommand{\tablescalefactor}{0.5}

\newsavebox{\mechanicallytablebox}
\savebox{\mechanicallytablebox}{\scalebox{\tablescalefactor}{\tikz[baseline=-1.6mm]\pic{mechanically};}}
\newcommand{\mechanicallytable}{\usebox{\mechanicallytablebox}}

\newsavebox{\polarimetricmechanicallytablebox}
\savebox{\polarimetricmechanicallytablebox}{\scalebox{\tablescalefactor}{\tikz[baseline=-1.6mm]\pic{polarimetricmechanically};}}
\newcommand{\polarimetricmechanicallytable}{\usebox{\polarimetricmechanicallytablebox}}

\newsavebox{\automotivetablebox}
\savebox{\automotivetablebox}{\scalebox{\tablescalefactor}{\tikz[baseline=-1.6mm]\pic{automotive};}}
\newcommand{\automotivetable}{\usebox{\automotivetablebox}}

\newsavebox{\polarimetricautomotivetablebox}
\savebox{\polarimetricautomotivetablebox}{\scalebox{\tablescalefactor}{\tikz[baseline=-1.6mm]\pic{polarimetricautomotive};}}
\newcommand{\polarimetricautomotivetable}{\usebox{\polarimetricautomotivetablebox}}

\newsavebox{\gprtablebox}
\savebox{\gprtablebox}{\scalebox{\tablescalefactor}{\tikz[baseline=-1.6mm]\pic{gpr};}}
\newcommand{\gprtable}{\usebox{\gprtablebox}}

\newsavebox{\cameratablebox}
\savebox{\cameratablebox}{\scalebox{\tablescalefactor}{\tikz[baseline=-1.6mm]\pic{camera};}}
\newcommand{\cameratable}{\usebox{\cameratablebox}}

\newsavebox{\lidartablebox}
\savebox{\lidartablebox}{\scalebox{\tablescalefactor}{\tikz[baseline=-1.6mm]\pic{lidar};}}
\newcommand{\lidartable}{\usebox{\lidartablebox}}

\newsavebox{\gyrotablebox}
\savebox{\gyrotablebox}{\scalebox{\tablescalefactor}{\tikz[baseline=-1.6mm]\pic{gyro};}}
\newcommand{\gyrotable}{\usebox{\gyrotablebox}}

\newsavebox{\accelerometertablebox}
\savebox{\accelerometertablebox}{\scalebox{\tablescalefactor}{\tikz[baseline=-1.6mm]\pic{accelerometer};}}
\newcommand{\accelerometertable}{\usebox{\accelerometertablebox}}

\newsavebox{\speedometertablebox}
\savebox{\speedometertablebox}{\scalebox{\tablescalefactor}{\tikz[baseline=-1.6mm]\pic{speedometer};}}
\newcommand{\speedometertable}{\usebox{\speedometertablebox}}

\newsavebox{\steeringangletablebox}
\savebox{\steeringangletablebox}{\scalebox{\tablescalefactor}{\tikz[baseline=-1.6mm]\pic{steeringangle};}}
\newcommand{\steeringangletable}{\usebox{\steeringangletablebox}}

\newsavebox{\gnsstablebox}
\savebox{\gnsstablebox}{\scalebox{\tablescalefactor}{\tikz[baseline=-1.6mm]\pic{gnss};}}
\newcommand{\gnsstable}{\usebox{\gnsstablebox}}

\newsavebox{\magnetometertablebox}
\savebox{\magnetometertablebox}{\scalebox{\tablescalefactor}{\tikz[baseline=-1.6mm]\pic{magnetometer};}}
\newcommand{\magnetometertable}{\usebox{\magnetometertablebox}}

\newsavebox{\crossmodaltablebox}
\savebox{\crossmodaltablebox}{\scalebox{\tablescalefactor}{\tikz[baseline=-1.6mm]\pic{crossmodal};}}
\newcommand{\crossmodaltable}{\usebox{\crossmodaltablebox}}

\newsavebox{\monomodaltablebox}
\savebox{\monomodaltablebox}{\scalebox{\tablescalefactor}{\tikz[baseline=-1.6mm]\pic{monomodal};}}
\newcommand{\monomodaltable}{\usebox{\monomodaltablebox}}

\newsavebox{\densetablebox}
\savebox{\densetablebox}{\scalebox{\tablescalefactor}{\tikz[baseline=-1.6mm]\pic{dense};}}
\newcommand{\densetable}{\usebox{\densetablebox}}

\newsavebox{\quasidensetablebox}
\savebox{\quasidensetablebox}{\scalebox{\tablescalefactor}{\tikz[baseline=-1.6mm]\pic{quasidense};}}
\newcommand{\quasidensetable}{\usebox{\quasidensetablebox}}

\newsavebox{\sparsetablebox}
\savebox{\sparsetablebox}{\scalebox{\tablescalefactor}{\tikz[baseline=-1.6mm]\pic{sparse};}}
\newcommand{\sparsetable}{\usebox{\sparsetablebox}}

\newsavebox{\crossmodalcaption}
\newsavebox{\monomodalcaption}
\newsavebox{\densecaption}
\newsavebox{\quasidensecaption}
\newsavebox{\sparsecaption}
\sbox\crossmodalcaption{\crossmodaltable}
\sbox\monomodalcaption{\monomodaltable}
\sbox\densecaption{\densetable}
\sbox\quasidensecaption{\quasidensetable}
\sbox\sparsecaption{\sparsetable}

\tikzset{invclip/.style={clip,insert path={{[reset cm]
        (-16383.99999pt,-16383.99999pt) rectangle (16383.99999pt,16383.99999pt)
      }}}}
\tikzset{
  pics/precfargridmapping/.style n args={2}{%
    code = {
      \begin{scope}[local bounding box=mappinggmscope]
      \path (0, 0) coordinate (gnssinput);
      \draw node [draw, below=0.75 * #1 of gnssinput, anchor=north west] (sr) {\makecell[c]{Static\\Reduction}};
      \path (sr.west) coordinate (radarinput);
      \draw [-latex] (gnssinput) -| (sr.north) coordinate [midway] (junction);
      \filldraw (junction) circle (0.03);
      \draw [-latex] (sr.east) -- ++(0.5 * #1, 0) node [draw, right] (ct) {\makecell[c]{Coordinate\\Trafo}};
      \draw [latex-] (ct.north) |- (gnssinput);
      \draw [-latex] (ct.east) -- ++(0.5 * #1, 0) node [draw, right] (cov) {\makecell[c]{Covariance\\Calculation}};
      \draw [-latex] (cov.east) -- ++(0.5 * #1, 0) node [draw, circle, right, inner sep=1pt] (sum) {\large +};
      \draw [latex-] (sum.north) |- ++(0.25 * #1, 0.5 * #1) node [draw, right] (delaymap) {T};
      \draw [latex-] (delaymap.east) -- ++(0.35 * #1, 0) coordinate (delaymapup) |- (sum.east);
      \draw (delaymapup |- sum.east) coordinate (sumfeedback);
      \filldraw (sumfeedback) circle (0.03);

      \path ($(cov.east)!0.5!(sum.west)$) -- ++(0, -0.5 * #1) coordinate (accumrectll);
      \path (sumfeedback |- delaymap.east) -- ++(0.15 * #1, 0.65 * #1) coordinate (accumrectur);
      \draw [densely dotted, thin] (accumrectll) rectangle (accumrectur);
      \draw (accumrectur) node [anchor=north east] {\tiny Accumulation};

      \path (mappinggmscope.north west) -- ++(-0.5 * #1, 0.25 * #1) coordinate (mapgmrectul);
      \path (mappinggmscope.south east) -- ++(0.35 * #1, -0.5 * #1) coordinate (mapgmrectlr);
      \draw [dashed, draw=#2] (mapgmrectul) rectangle (mapgmrectlr);

      \draw (mapgmrectlr) node [anchor=south east] {\scriptsize PreCFAR
        Covariance Gridmapping (\ref{sec:gridmap})};
    \end{scope}
    }},
  pics/landmarkdetector/.style n args={2}{%
    code = {
    \begin{scope}[local bounding box=mappingdetectorscope]
      \path (0, 0) -- ++(0.75 * #1, 0) coordinate (detmiddle);
      \node (pcfarnode) [above right=0.15 * #1 and 0.1 * #1 cm of detmiddle, draw] {\makecell[c]{pCFAR\\Detector}};
      \node (ridgenode) [below right=0.15 * #1 and 0.1 * #1 cm of detmiddle, draw] {\makecell[c]{Ridge\\Detector}};
      \filldraw (0, 0) circle (0.03);
      \draw [-latex] (0, 0) |- (pcfarnode.west);
      \draw [-latex] (0, 0) |- (ridgenode.west);

      \path (mappingdetectorscope.north west) -- ++(-0.5 * #1, 0.25 * #1) coordinate (mapdetectorrectul);
      \path (mappingdetectorscope.south east) -- ++(0.5 * #1, -0.5 * #1) coordinate (mapdetectorrectlr);
      \draw [dashed, draw=#2] (mapdetectorrectul) rectangle (mapdetectorrectlr);
      \draw (mapdetectorrectlr) node [anchor=south east] {\scriptsize Landmark
        Detection (\ref{sec:lm_extraction})};
    \end{scope}
  }}
}

\tikzset{
  vertical shaded border/.style args={#1 and #2}{
    append after command={
      \pgfextra{%
        \begin{pgfinterruptpath}
          \path[rounded corners,top color=#1,bottom color=#2]
          ($(\tikzlastnode.south west)+(-2\pgflinewidth,-2\pgflinewidth)$)
          rectangle
          ($(\tikzlastnode.north east)+(2\pgflinewidth,2\pgflinewidth)$);
        \end{pgfinterruptpath}
      }
    }
  }
}

\tikzset{
  hyperref node/.style={
    alias=sourcenode,
    append after command={
      let     \p1 = (sourcenode.north west),
      \p2=(sourcenode.south east),
      \n1={\x2-\x1},
      \n2={\y1-\y2} in
      node [inner sep=0pt, outer sep=0pt,anchor=north west,at=(\p1)] {\hyperref[#1]{\XeTeXLinkBox{\phantom{\rule{\n1}{\n2}}}}}
    }
  }
}

\makeatletter
\tikzset{
  hyperref rect/.style={
    path picture={%
      \pgfpointanchor{path picture bounding box}{south west}%
      \pgf@xb-\pgf@x
      \pgf@yb-\pgf@y
      \pgfpointanchor{path picture bounding box}{north east}%
      \advance\pgf@xb\pgf@x
      \advance\pgf@yb\pgf@y
      \pgftext[at={\pgfqpoint{\pgf@x}{\pgf@y}},right,top]{\hyperref[#1]{\vrule height\pgf@yb depth0ptwidth0pt\vrule height0ptdepth0ptwidth\pgf@xb}}%
    }
  }
}
\makeatother

\newcommand\clipdraw[2]{%
      \begin{scope}
        \clip #1;
        \draw [#2] #1;
      \end{scope}}

\tikzset{
  pics/coloredhalfcube/.style n args={4}{%
    code = {
      \clipdraw{
        (#1, #1, #1) --
        (#1,  0, #1) --
        (#1,  0,  0) --
        (#1, #1,  0) -- cycle}
        {draw=#4, fill=#3, opacity=#2};

      \clipdraw{
        ( 0,  0,  0) --
        ( 0,  0, #1) --
        (#1,  0, #1) --
        (#1,  0,  0) -- cycle}
        {draw=#4, fill=#3, opacity=#2};

      \clipdraw{
        (#1, #1, #1) --
        (#1,  0, #1) --
        ( 0,  0, #1) --
        ( 0, #1, #1) -- cycle}
        {draw=#4, fill=#3, opacity=#2};
     }},
    pics/coloredhalfcubeindependentscale/.style n args={6}{%
    code = {
      \clipdraw{
        (#1, #2, #3) --
        (#1,  0, #3) --
        (#1,  0,  0) --
        (#1, #2,  0) -- cycle}
        {draw=#6, fill=#5, opacity=#4};

      \clipdraw{
        ( 0,  0,  0) --
        ( 0,  0, #3) --
        (#1,  0, #3) --
        (#1,  0,  0) -- cycle}
        {draw=#6, fill=#5, opacity=#4};

      \clipdraw{
        (#1, #2, #3) --
        (#1,  0, #3) --
        ( 0,  0, #3) --
        ( 0, #2, #3) -- cycle}
        {draw=#6, fill=#5, opacity=#4};
    }},
  pics/coloredcube/.style n args={4}{%
    code = {
      \clipdraw{
        ( 0,  0,  0) --
        (#1,  0,  0) --
        (#1, #1,  0) --
        ( 0, #1,  0) -- cycle}
        {draw=#4, fill=#3, opacity=#2};

      \clipdraw{
        ( 0,  0,  0) --
        ( 0,  0, #1) --
        ( 0, #1, #1) --
        ( 0, #1,  0) -- cycle}
        {draw=#4, fill=#3, opacity=#2};

      \clipdraw{
        (#1, #1, #1) --
        ( 0, #1, #1) --
        ( 0, #1,  0) --
        (#1, #1,  0) -- cycle}
        {draw=#4, fill=#3, opacity=#2};

      \draw (0, 0, 0) pic {coloredhalfcube={#1}{#2}{#3}{#4}};
      }},
  pics/coloredcubes/.style n args={7}{%
  code = {
    \foreach \z in {1, ..., #3}
    {
      \foreach \y in {1, ..., #2}
      {
        \foreach \x in {1, ..., #1}
        {
          \draw (\x * #4 - #4, \y * #4 - #4, \z * #4 - #4) pic {coloredcube={#4}{#5}{#6}{#7}};
        }
      }
    }
  }},
  pics/coloredhalfcubes/.style n args={7}{%
  code = {
    \foreach \z in {1, ..., #3}
    {
      \foreach \x in {1, ..., #1}
      {
        \draw (\x * #4 - #4, 0, \z * #4 - #4) pic {coloredhalfcube={#4}{#5}{#6}{#7}};
      }
    }

    \foreach \z in {1, ..., #3}
    {
      \foreach \y in {#2, ..., 1}
      {
        \draw (#1 * #4 - #4, \y * #4 - #4, \z * #4 - #4) pic {coloredhalfcube={#4}{#5}{#6}{#7}};
      }
    }

    \foreach \x in {1, ..., #1}
    {
      \foreach \y in {#2, ..., 1}
      {
        \draw (\x * #4 - #4, \y * #4 - #4, #3 * #4 - #4) pic {coloredhalfcube={#4}{#5}{#6}{#7}};
      }
    }
  }},
  pics/coloredhalfcubesmulti/.style n args={5}{%
  code = {
    \foreach \z in {1, ..., #3}
    {
      \foreach \x in {1, ..., #1}
      {
        \draw (\x * #4 - #4, 0.5 * #4, \z * #4 - #4) pic {coloredhalfcubeindependentscale={#4 * 0.5}{#4 * 0.5}{#4}{#5}{Paired-E}{Paired-F}};
        \draw (\x * #4 - #4 + 0.5 * #4, 0.5 * #4, \z * #4 - #4) pic {coloredhalfcubeindependentscale={#4 * 0.5}{#4 * 0.5}{#4}{#5}{Paired-G}{Paired-H}};
        \draw (\x * #4 - #4, 0, \z * #4 - #4) pic {coloredhalfcubeindependentscale={#4 * 0.5}{#4 * 0.5}{#4}{#5}{Paired-A}{Paired-B}};
        \draw (\x * #4 - #4 + 0.5 * #4, 0, \z * #4 - #4) pic {coloredhalfcubeindependentscale={#4 * 0.5}{#4 * 0.5}{#4}{#5}{Paired-C}{Paired-D}};
      }
    }

    \foreach \z in {1, ..., #3}
    {
      \foreach \y in {#2, ..., 1}
      {
        \draw (#1 * #4 - #4, \y * #4 - #4 + 0.5 * #4, \z * #4 - #4) pic {coloredhalfcubeindependentscale={#4 * 0.5}{#4 * 0.5}{#4}{#5}{Paired-E}{Paired-F}};
        \draw (#1 * #4 - #4 + 0.5 * #4, \y * #4 - #4 + 0.5 * #4, \z * #4 - #4) pic {coloredhalfcubeindependentscale={#4 * 0.5}{#4 * 0.5}{#4}{#5}{Paired-G}{Paired-H}};
        \draw (#1 * #4 - #4, \y * #4 - #4, \z * #4 - #4) pic {coloredhalfcubeindependentscale={#4 * 0.5}{#4 * 0.5}{#4}{#5}{Paired-A}{Paired-B}};
        \draw (#1 * #4 - #4 + 0.5 * #4, \y * #4 - #4, \z * #4 - #4) pic {coloredhalfcubeindependentscale={#4 * 0.5}{#4 * 0.5}{#4}{#5}{Paired-C}{Paired-D}};
      }
    }

    \foreach \x in {1, ..., #1}
    {
      \foreach \y in {#2, ..., 1}
      {
        \draw (\x * #4 - #4, \y * #4 - #4 + 0.5 * #4, #3 * #4 - #4) pic {coloredhalfcubeindependentscale={#4 * 0.5}{#4 * 0.5}{#4}{#5}{Paired-E}{Paired-F}};
        \draw (\x * #4 - #4 + 0.5 * #4, \y * #4 - #4 + 0.5 * #4, #3 * #4 - #4) pic {coloredhalfcubeindependentscale={#4 * 0.5}{#4 * 0.5}{#4}{#5}{Paired-G}{Paired-H}};
        \draw (\x * #4 - #4, \y * #4 - #4, #3 * #4 - #4) pic {coloredhalfcubeindependentscale={#4 * 0.5}{#4 * 0.5}{#4}{#5}{Paired-A}{Paired-B}};
        \draw (\x * #4 - #4 + 0.5 * #4, \y * #4 - #4, #3 * #4 - #4) pic {coloredhalfcubeindependentscale={#4 * 0.5}{#4 * 0.5}{#4}{#5}{Paired-C}{Paired-D}};
      }
    }
  }}
}

\tikzset{
  posenode/.style={
    draw,
    single arrow,
    single arrow head extend=0,
    shape border uses incircle,
    inner sep=0pt
  }
}

\definecolor{orcidlogocol}{HTML}{A6CE39}
\tikzset{
  orcidlogo/.pic={
    \fill[orcidlogocol] svg{M256,128c0,70.7-57.3,128-128,128C57.3,256,0,198.7,0,128C0,57.3,57.3,0,128,0C198.7,0,256,57.3,256,128z};
    \fill[white] svg{M86.3,186.2H70.9V79.1h15.4v48.4V186.2z}
    svg{M108.9,79.1h41.6c39.6,0,57,28.3,57,53.6c0,27.5-21.5,53.6-56.8,53.6h-41.8V79.1z M124.3,172.4h24.5c34.9,0,42.9-26.5,42.9-39.7c0-21.5-13.7-39.7-43.7-39.7h-23.7V172.4z}
    svg{M88.7,56.8c0,5.5-4.5,10.1-10.1,10.1c-5.6,0-10.1-4.6-10.1-10.1c0-5.6,4.5-10.1,10.1-10.1C84.2,46.7,88.7,51.3,88.7,56.8z};
  }
}

\newcommand\orcidicon[1]{\href{https://orcid.org/#1}{\mbox{\scalerel*{\tikzexternaldisable\begin{tikzpicture}[yscale=-1,transform shape] \pic{orcidlogo}; \end{tikzpicture} }{|}}}}

\title{Landmark-based Vehicle Self-Localization Using Automotive Polarimetric Radars}

\author{ %
  Fabio~Weishaupt~\orcidicon{0000-0001-6489-0522}, %
  Julius~F.~Tilly~\orcidicon{0000-0001-9659-3389}, %
  Nils~Appenrodt~\orcidicon{0000-0002-1053-9136}, %
  Pascal~Fischer~\orcidicon{0000-0002-7861-0226}, %
  J\"urgen~Dickmann~\orcidicon{0000-0002-4328-3368} %
  and~Dirk~Heberling~\orcidicon{0000-0003-2438-9607},~\IEEEmembership{Senior Member, IEEE} %
  \thanks{\noindent Manuscript received August 13, 2023; revised March 2, 2024. Date of publication %
    ZZZ. This work results from the cooperative project @CITY \textendash{} Automated Cars %
    and Intelligent Traffic in the City \textendash{} and is supported by the Federal Ministry %
    for Economic Affairs and Energy on the basis of a decision by the German %
    Bundestag. The Associate Editor for this paper was XXX. \textit{(Corresponding author: %
    Fabio Weishaupt.)}} %
  \thanks{Fabio Weishaupt, Nils Appenrodt, Julius F. Tilly and J\"urgen Dickmann are %
    with Research \& Development at Mercedes-Benz AG, 70565 Stuttgart, Germany (e-mail: %
    \href{mailto:fabio.weishaupt@mercedes-benz.com}{fabio.weishaupt@mercedes-benz.com}).} %
  \thanks{Pascal Fischer was with Research \& Development at Mercedes-Benz AG, 70565 %
    Stuttgart, Germany.} %
  \thanks{Dirk Heberling is with the Institute of High Frequency Technology, RWTH %
    Aachen University, 52074 Aachen, Germany, and with the Fraunhofer %
    Institute for High Frequency Physics and Radar Techniques FHR, 53343 %
    Wachtberg, Germany.} %
}

\markboth{WEISHAUPT \MakeLowercase{\textit{et al.}}: LANDMARK-BASED
  VEHICLE SELF-LOCALIZATION USING AUTOMOTIVE POLARIMETRIC RADARS}%
{WEISHAUPT \MakeLowercase{\textit{et al.}}: LANDMARK-BASED
  VEHICLE SELF-LOCALIZATION USING AUTOMOTIVE POLARIMETRIC RADARS}

\maketitle
\copyrightnotice

\begin{abstract}
  Automotive self-localization is an essential task for any automated driving
  function. This means that the vehicle has to reliably know its position and
  orientation with an accuracy of a few centimeters and degrees, respectively.
  This paper presents a radar-based approach to self-localization, which
  exploits fully polarimetric scattering information for robust landmark
  detection. The proposed method requires no input from sensors other than
  radar during localization for a given map. By association of landmark
  observations with map landmarks, the vehicle's position is inferred. Abstract
  point- and line-shaped landmarks allow for compact map sizes and, in
  combination with the factor graph formulation used, for an efficient
  implementation.

  Evaluation of extensive real-world experiments in diverse environments shows
  a promising overall localization performance of \SI{0.12}{\meter} RMS absolute
  trajectory and \SI{0.43}{\degree} RMS heading error by leveraging the
  polarimetric information. A comparison of the performance of different levels
  of polarimetric information proves the advantage in challenging scenarios.
\end{abstract}


\section{INTRODUCTION}
\begin{figure}[t]
  \resizebox{\columnwidth}{!}{{\input{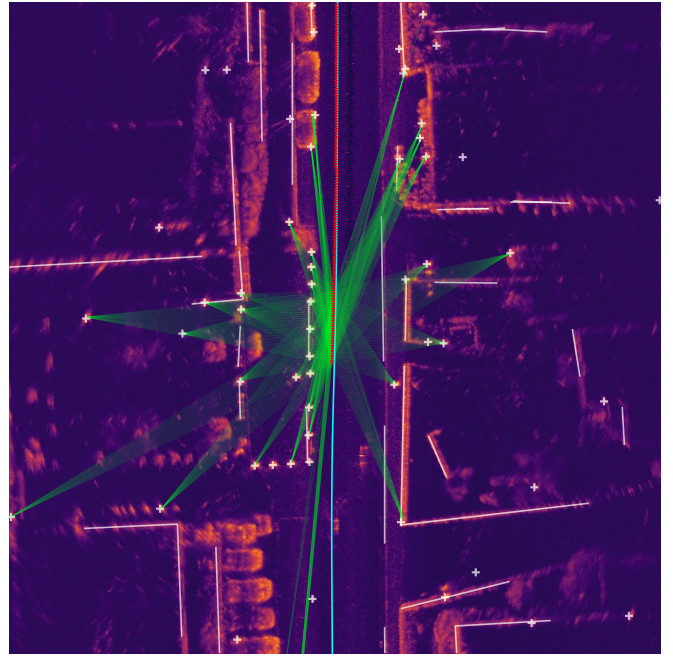}}}
  \caption{The proposed localization framework estimates the ego-vehicle's pose
    by optimizing  a sliding window pose graph. An association (green) is shown
    for each pose from which a map landmark (white) is observed. The resulting
    localization trajectory is shown in red, the ground truth trajectory in
    cyan.}
  \label{fig:appetizer}
\end{figure}
\IEEEPARstart{W}{hile} for lower level advanced driver assistance systems radar
sensors mainly focused on other road users, this is no longer sufficient for
automated driving functions. One reason for this is the requirement that the
vehicle must know its position and orientation (pose) very precisely and with
high reliability, e.g. for path planning. The availability and precision of
classical global navigation satellite systems (GNSS) are inadequate. Although
differential GNSS approaches might meet the precision requirement, they are not
available in scenarios without an unobstructed view to the satellites, e.g. in
tunnels or between high-rise buildings. Therefore, exteroceptive sensors are
used to perceive the static environment in order to derive pose estimates from
that~\cite{lu_loc_overview}. For robustness, multiple sensor modalities are
typically combined, with one of them often being radar. While camera and lidar
benefit from higher angular resolution, they suffer from a lack of direct range
measuring capability and high cost, respectively. In contrast, radar has
moderate costs and is a modality that has been integrated into serial products
for decades. Furthermore, radar is robust to inclement weather conditions such
as rain, fog and snow due to a comparatively long wavelength~\cite{bijelic_fog}.
Even though first experimental frequency modulated continuous wave lidar
prototypes are being built, radar is still the only serially produced long-range
sensing modality capable of measuring relative velocity instantaneously. This is
desirable for localization applications, as one can easily distinguish between
other moving road users and the static environment~\cite{holder_slam}.

In order to derive meaningful pose estimates from exteroceptive sensor data, a
map containing a priori knowledge about objects expected in a given environment,
including their positions, is required (cf.~Fig.~\ref{fig:appetizer}). If the
map objects are georeferenced, a relative localization within this map
corresponds to global localization. The representation of the map elements can
vary considerably, ranging from a raw collection of measurements from a previous
mapping pass, over a grid-based representation, to landmarks. The memory size
requirements decrease in the given order, so only the landmark-based approach is
considered feasible for large-scale nationwide applications. Typical landmarks
suitable for radar include point-shaped objects such as tree trunks, guideposts,
traffic sign and street lamp poles, as well as line-shaped areas like
curbstones, buildings and fences~\cite{werber_interesting_areas}.

Polarimetric radar is an established technique in space- and airborne
remote sensing applications for several decades~\cite{lee_polsar}. By
transmitting and receiving different polarizations, additional information about
the scattering mechanisms that occurred during signal propagation can be
derived. A classic example for airborne systems is the polarimetry-based
distinction between single-bounce scattering of the sea and double-bounce
scattering of buildings in urban environments. Although the measurement
setup and operating frequency of airborne radars are significantly different
from ground-based automotive applications, the fundamental principle of
scattering parity can be transferred. Simple single-bounce objects such as poles
of traffic signs or street lamps fall into the category of odd-bounce
scattering, whereas door or window recesses of buildings and guardrail posts
are typical examples of even-bounce scattering~\cite{iqbal_pol_analysis,
weishaupt_pol_analysis, weishaupt_covariance}. With the everlasting desire for
an increasing number of channels in order to further improve angle finding in
automotive radars, highly integrated circuits (ICs) became available in recent
years such that three-digit channel counts can be realized by application of the
multiple-input multiple-output (MIMO) principle in form factors that are
feasible for passenger vehicle packaging. Therefore, the disadvantage of needing
more channels for polarimetry is no longer prohibitive. While one has to accept
some disadvantages in angle finding when available channels are spent for
polarimetry, the number of remaining channels is still sufficient for an usable
angle finding due to the initial large number with the latest ICs. Moreover,
angular resolution and sidelobe levels, which would benefit from high channel
counts, are typically not performance limiting factors for localization
applications. Due to the assumption of a moving ego-vehicle, the static
surroundings can be resolved by Doppler.

This publication proposes a complete landmark-based localization framework that
exploits polarimetric information by identifying and localizing landmarks more
reliably and precisely in comparison to non-polarimetric approaches.
Based on the introduced polarimetric representation of the static environment,
the capability to distinguish between different scattering mechanisms is used to
improve the localization accuracy without the requirement to store polarimetric
information in the map. By omitting the latter, the proposed approach does not
depend by principle on a mapping vehicle with a polarimetric radar avoiding a
potentially prohibitive disadvantage. The measurement setup used is fully
polarimetric, i.e. the full complex scattering matrix is available. By selecting
different subsets of the full matrix, the benefit of different polarimetric
implementations of such a system can be compared, i.e. whether a
dual-polarimetric system capable of discriminating between even- and odd-bounce
scattering already provides the pursued key advantage.

The main contributions of this article are:

\begin{itemize}
  \item Adaption of a radar-based ego-motion estimation approach to
    counteract a typical drawback of polarimetry,
  \item Scalable static environment representation created from low-level
    polarimetric data,
  \item Polarimetry-leveraging radar-only localization framework which is
    based on point- and line-shaped landmark maps,
  \item Evaluation of real-world localization performance using different levels
    of polarimetric information.
\end{itemize}

The paper is structured as follows. Section
\ref{sec:related_work} provides a comprehensive review of literature related to
finding the ego-vehicle's pose and related to polarimetric radar
in automotive contexts. The proposed polarimetric, landmark-based localization
approach is described in detail in Section \ref{sec:proposed_approach}. Based on
the description of the experimental vehicle setup and the evaluation framework,
the results of the experiments are presented and discussed in Section
\ref{sec:experimental_evaluation}. Finally, Section \ref{sec:conclusion}
concludes the article and gives an outlook on possible future research
directions.

\section{RELATED WORK}
\label{sec:related_work}
\noindent A sizeable body of literature exists on the vehicle self-localization
using exteroceptive sensors. The majority of work is based on lidar or camera
sensors with a recent increase in interest in using radar. Due to the radar
focus of this work, the other sensor modalities are not covered in this section.
For this, the reader is referred to a recent review paper
\cite{lu_loc_overview}.
The research area related to the estimation of the ego-vehicle's pose using
radar sensors includes several sub-fields: simultaneous localization and mapping
(SLAM), place recognition, ego-motion estimation and map-based
self-localization.

SLAM is an approach to solve mapping and trajectory estimation in unknown
environments at the same time. While such techniques are interesting for
creating the a priori map for map-based self-localization, the achieved
accuracies on longer routes and the loop closure requirement for improved
accuracy~\cite{holder_slam, hong_slam} renders SLAM insufficient as the sole
source of pose information in automated driving applications.
Although place recognition is also a technique for estimation of the
ego-vehicle's location, it is not comparable to finding consistent sub-meter
accurate trajectories because errors within a few meters are considered a
successful localization~\cite{saftescu_pr, wang_pr}.

Therefore, SLAM and place recognition objectives are too different from this
paper's focus such that only the literature on the two remaining sub-fields is
considered in more detail in the following. Because polarimetry is still
relatively unexplored in the automotive radar community, a short overview of the
research directions is given and the preparatory work for this paper is
presented.

\subsection{Ego-motion estimation}
\label{sec:related_work_egomotion}
\noindent Ego-motion estimation %
is a technique to derive the parameters of motion that the ego-vehicle is
experiencing (usually translational and angular velocities). By integrating
the estimates over time, an ego-pose can be calculated. However, the result
will increasingly drift with longer integration times, so that using it alone is
not sufficient for automated driving. Nevertheless, ego-motion estimation is
often part of a localization system where the drift error is corrected by map
correlation.
For radar-based ego-motion estimation, two different approaches are common,
depending on the type of radar used. Mechanically scanning
$\SI{360}{\degree}$ radars, such as those used in the public ``Oxford Radar
RobotCar Dataset''~\cite{barnes_dataset}
and ``MulRan Dataset for Urban Place Recognition''~\cite{giseop_dataset}
dataset, have no Doppler measurement capability, so two consecutive scans must
be matched. This typically involves finding and matching keypoints between the
scans and is still subject of research with both classical~\cite{cen_egomotion,
  cen_egomotion_2} as well as learned approaches~\cite{barnes_egomotion,
  burnett_egomotion}.
For common automotive radars with a beamforming-based angular and Doppler-based
velocity measuring capability, the motion state can be estimated instantaneously
from a single scan by exploiting the dependence of the measured radial velocity
on the azimuth angle for static infrastructure~\cite{kellner_egomotion}.

\subsection{Radar-based self-localization with maps}

\begin{table}[b]
  \renewcommand{\arraystretch}{1.1}
  \caption{Translation of sensor and map type symbols}
  \label{tab:symbol_tanslation}
  \setlength{\tabcolsep}{4pt}
  \centering
  \begin{tabularx}{\columnwidth}{p{.3cm}X|p{.3cm}l||p{.3cm}l}
    \toprule
    \multicolumn{4}{c||}{\bfseries \footnotesize Sensor Types} &
                                                          \multicolumn{2}{c}{\bfseries\footnotesize Map Types}\\
    \midrule
    \automotivetable & Beamforming radar & \gyrotable & Gyroscope & \crossmodaltable{} & cross-modal\\
    \mechanicallytable & Scanning radar & \accelerometertable & Accelerometer & \monomodaltable{} & mono-modal\\
    \gprtable & Ground penetrating radar & \speedometertable & Speedometer & \sparsetable{} & sparse\\
    \, + & Polarimetric radar & \steeringangletable & Steering angle & \quasidensetable{} & quasi-dense\\
    \lidartable & Lidar & \gnsstable & GNSS & \densetable{} & dense\\
    \cameratable & Camera & \magnetometertable & Magnetometer\\
    \bottomrule
  \end{tabularx}
\end{table}

\begin{table*}
  \renewcommand{\arraystretch}{1.27}
  \caption{Overview of approaches to map-based self-localization using radar
    sensors. Resulting localization errors are given as RMSE values unless
    otherwise noted. The symbol ${}^\dagger$ indicates that \SI{68}{\percent} of
    errors were below the given value. The symbol * denotes MAE values, ${}^\#$
    marks indirect values, e.g. estimates from graphs.}
  \label{tab:self_localization_overview}
  \setlength{\tabcolsep}{15pt}
  \centering
  \begin{tabularx}{\textwidth}{@{}Xclr@{\hspace{0.1cm}}l@{\hspace{0.2cm}}crZc@{\hspace{0.2cm}}cZc@{\hspace{0.2cm}}cZ@{}}
    \toprule
    {\bfseries Author} & {\bfseries Reference} & {\bfseries Year} &
    \multicolumn{3}{c}{\bfseries Sensor Set} & {\bfseries Route} &
    {\bfseries Environment} & \multicolumn{3}{c}{\bfseries Map} &
    \multicolumn{3}{c}{\bfseries Error (\si{\meter})}\\
    & & & \multicolumn{2}{c}{Radar} & additional & & & Modality & Sparsity & & longitudinal & lateral & phi\\
    \midrule
    Clark et al. & \cite{clark_loc} & 1998 & 1x & \polarimetricmechanicallytable & \speedometertable{}
     \steeringangletable & \SI{0.2}{\kilo\meter} & one test site &
     \crossmodaltable{} & \sparsetable & surveyed artificial beacon positions & \textendash{} & \textendash{} &\\
    Lundgren et al. & \cite{lundgren_loc} & 2014 & 1x & \automotivetable &
    \speedometertable{} \gyrotable{} \gnsstable{}
     & \SI{5}{\kilo\meter} & only rural road & \monomodaltable{} & \sparsetable & points (radar) and lines (camera)
     (no camera is also evaled)& $\approx 0.50^{\#}$& $\approx 0.50^{\#}$ & \\
    Rapp et al. & \cite{rapp_loc} & 2015 & 4x & \automotivetable & \speedometertable{} \gyrotable
    & \SI{0.1}{\kilo\meter} & 11 map building one loc & \monomodaltable{} & \densetable & 0.1m grid size gridmap & \multicolumn{2}{c}{${0.16}^{\dagger, \#}$} & \\
    Cornick et al. & \cite{cornick_loc} & 2016 & 1x & \gprtable & \accelerometertable{}
    \gyrotable{} \gnsstable{} & \SI{1.6}{\kilo\meter} %
    & additional test with minor losses of
    loc & \monomodaltable{} & \densetable & volumetric soil map & 0.06 & 0.04 & -\\
    Schuster et al. & \cite{schuster_loc} & 2016 & 4x & \automotivetable & \speedometertable{} \gyrotable{}
     & \SI{0.4}{\kilo\meter} & multiple tests (parking lot, different time) &
     \monomodaltable{} & \quasidensetable & radar clusters (200 kB), same SSU &
    \multicolumn{2}{c}{$> 1.0^{\#}$} & ? \\
    Ward et al. & \cite{ward_loc} & 2016 & 2x & \automotivetable & \speedometertable{} \gyrotable{} & %
    \SI{5.4}{\kilo\meter} & one drive (but HW and residential) & \monomodaltable{} & \densetable &
    full pcl & 0.38 & 0.07 & \\
    Yoneda et al. & \cite{yoneda_loc} & 2018 & 9x & \automotivetable & \speedometertable{} \gyrotable{}
     & \SI{2.2}{\kilo\meter} & inkl snow & \monomodaltable{} & \densetable & image (correlation) & 0.19 & 0.26 & \\
    Li et al. & \cite{li_loc} & 2019 & 1x & \automotivetable & \textendash{} & \SI{0.2}{\kilo\meter} & one test (urban)
    & \monomodaltable{} & \quasidensetable & point RO & \multicolumn{2}{c}{0.65} & ? \\
    Iannucci et al. & \cite{iannucci_loc} & 2020 & 3x & \automotivetable & \speedometertable{} \gyrotable{} & \SI{1.5}{\hour} &
    urban & \crossmodaltable{} & \quasidensetable & SIFT correspondence points & \multicolumn{2}{c}{${\approx 0.3}^{\dagger, \#}$} & \\
    J\"urgens et al. & \cite{juergens_localization} & 2020 & 6x & \automotivetable & \textendash &
    \SI{2}{\kilo\meter} & urban/industrial & \crossmodaltable{} & \sparsetable & semantic HD map & $0.20^{\dagger, \#}$ & $0.25^\dagger$ & -\\
    Narula et al. & \cite{narula_localization} & 2020 & 3x & \automotivetable & \speedometertable{} \gyrotable{} &
    \SI{1.5}{\hour} & urban & \monomodaltable{} & \densetable & full pcl & \multicolumn{2}{c}{${\approx 0.3}^{\dagger,\#}$} & \\
    Ort et al. & \cite{ort_loc} & 2020 & 1x & \gprtable & \speedometertable{} \gyrotable{}
    \magnetometertable{} & \SI{7}{\kilo\meter} & rural bad weather
    included & \monomodaltable{} & \densetable & volumetric soil map (160GB) & 0.17* & 0.26* & -\\
    Pishehvari et al. & \cite{pishehvari_loc} & 2020 & 1x & \automotivetable & \speedometertable{} \gyrotable{} &
    \SI{0.2}{\kilo\meter} & two tests (urban) & \crossmodaltable{} & \sparsetable &
    RD reg (lm map) & \multicolumn{2}{c}{0.16} & 0.02° \\
    Yin et al. & \cite{yin_gan_loc} & 2020  & 1x & \mechanicallytable & \textendash{} & \SI{9}{\kilo\meter} %
    & multiple tests (4) & \crossmodaltable{} & \densetable & lidar map BEV image & \multicolumn{2}{c}{$>$ 6.0} & $>$ 2\\
    Engel et al. & \cite{engel_loc} & 2021 & 2x & \automotivetable & \lidartable{} \cameratable & \SI{6}{\kilo\meter} & one
          drive only & \monomodaltable{} & \sparsetable & landmarks & 0.18 & 0.16 & 0.6\\
    Otake & \cite{otake_loc} & 2021 & 5x & \automotivetable & \speedometertable{} \gyrotable{} & \SI{1.4}{\kilo\meter} & industrial &
    \monomodaltable{} & \densetable & one drive only (slow speed) & 0.07 & 0.07 & 0.4 \\
    Burnett et al. & \cite{burnett_loc} & 2022 & 1x & \mechanicallytable & \textendash{} & %
    \SI{8}{\kilo\meter} %
    & all-weather (snow) but urban & \monomodaltable{} & \quasidensetable & filtered PCL (bfar, no LMs) &
    0.13 & 0.12 & 0.22\\
    Yanase et al. & \cite{yanase_radar_lidar_loc} & 2022 & 9x & \automotivetable & \speedometertable{}\gyrotable{}& %
     \SI{4.6}{\kilo\meter} & inkl snow & \monomodaltable{} & \densetable & image (correlation) & 0.16 & 0.14 & \\
    Yin et al. & \cite{yin_loc} & 2022 & 1x & \mechanicallytable & \textendash{} & \SI{9}{\kilo\meter} %
    & multiple tests
    different countries & \crossmodaltable{} & \densetable & lidar map BEV image & \multicolumn{2}{c}{$\approx$ 1.0} & 1.5\\
    \midrule
    \multicolumn{2}{c}{\bfseries This work} & 2022 &  3x & \polarimetricautomotivetable & \textendash & \SI{10}{\kilo\meter} &
    multiple drives & \monomodaltable{} & \sparsetable & same sensor & 0.11 & 0.06 & \\
    \addlinespace
    \bottomrule
  \end{tabularx}
\end{table*}

\noindent This section provides an overview of work related to finding the
ego-vehicle's pose relative to an a priori map by using radar. Because the
amount of research in this exact field is still limited, it is immature in a
sense that no gold standard approach has been established yet. Therefore, this
overview is broad and includes approaches that in part differ greatly. It is
intended to help the reader put this work into context.
Table~\ref{tab:self_localization_overview} accompanies this section for a clear
and concise comparison.

One of the most important differentiating factors between the approaches is the
used sensor setup during the localization phase. The type of radar and any
additional supportive sensor modality both play a crucial role for the system's
design. Table~\ref{tab:symbol_tanslation} introduces pictograms for brevity in
Table~\ref{tab:self_localization_overview}.
Accordingly, each work is classified into one of the following three categories
of radar: {\itshape (automotive) beamforming}, {\itshape (mechanically)
  scanning} or {\itshape ground penetrating radar}. The optional + symbol
indicates that the used radar is polarimetric.

The former two scan the vehicle's above-ground environment and are therefore
suitable for advanced driver assistance system functions such as adaptive cruise
control or blind spot monitoring in addition to localization. In contrast, ground
penetrating radars only scan for sub-surface features and are limited to
localization applications~\cite{cornick_loc, ort_loc}. The distinction between
the two radar types (beamforming \& mechanically scanning) perceiving the
above-ground environment was already introduced in
\ref{sec:related_work_egomotion}. Further considerations for or against one of
these radar types include the number of sensors needed for omni-directional
coverage, ease of integration and packaging, vehicle design, power consumption,
update rate and wear~\cite{kung_automotive_scanning}.

As far as the additional sensing modalities are concerned, most approaches rely
on motion information from proprioceptive sources such as wheel tick and
gyroscope sensors. These measurements typically serve as an initial pose
estimate and are subsequently corrected for any drift by the exteroceptive sensor
measurements. Only a minority of approaches is radar-only. They derive the
odometry from the radar alone, either through scan matching~\cite{li_loc,
  yin_gan_loc, burnett_loc, yin_loc} or instantaneous Doppler-based estimation
\cite{juergens_localization}. When different combinations of sensor modalities
are evaluated in the paper, the minimal set that includes radar and the
corresponding localization error is chosen for better comparability in
Table~\ref{tab:self_localization_overview}.
Due to the different research directions of the considered papers, the route
lengths for error evaluation differ considerably. However, in order to provide a
comprehensive overview to the reader, the wide variety of approaches is still
included. If different routes are used for evaluation or if multiple passes of
the same route are evaluated, Table~\ref{tab:self_localization_overview} lists
the longest unique route.

Another distinctive property of each localization approach is the map type. The
general categories for classification are: {\itshape sparse} versus
{\itshape dense} and {\itshape cross-modal} versus {\itshape mono-modal} (for
pictograms see Table~\ref{tab:symbol_tanslation}).
The former differentiation describes the density of landmarks in the map
and serves as an indicator for the required memory size to store the map and
consequentially the scalability for large-scale deployments.
Cross-modal maps are acquired with a sensor modality other than the one used
during the localization phase, whereas mono-modal maps are usually constructed
from measurements with the same sensor setup that is intended for localization.

Sparse, cross-modal map representations include manually surveyed positions
of artificial beacons~\cite{clark_loc} as well as pole-like objects in semantic
high definition (HD) maps~\cite{juergens_localization, pishehvari_loc}. In
contrast, sparse, mono-modal maps are the result of feature extraction
algorithms applied to measurement data of the same sensors used during
localization \cite{lundgren_loc, engel_loc}.
The advantages of desirable scaling properties of sparse maps are accompanied by
a more challenging association between online measurements and map elements
as well as a higher probability of ambiguities when compared to dense maps. With
regard to dense, mono-modal maps, the density covers a wide range, so that a
further distinction between quasi-dense and dense is made in
Table~\ref{tab:self_localization_overview}. It ranges from a point cloud
clustered over multiple measurement cycles~\cite{schuster_loc}, over a simple
accumulation of unfiltered~\cite{narula_localization, ward_loc} and
filtered~\cite{burnett_loc} point clouds as well as image-like two-dimensional
grids~\cite{yoneda_loc, otake_loc, yanase_radar_lidar_loc}, to volumetric grids
for the ground penetrating systems \cite{cornick_loc, ort_loc}. Large-scale
deployments of ground penetrating localization are challenging not only due to
the large memory footprint, but also because each lane of a road must be mapped
individually. For all other approaches, a single pass is conceptually
sufficient. The remaining dense, cross-modal approaches take advantage of better
availability of camera- and lidar-based maps compared to radar. These range from
map matching techniques based on density map correlation of visual SIFT features
with radar detection density maps~\cite{iannucci_loc}, over the transformation
of radar images into lidar representations using a generative adversarial
network~\cite{yin_gan_loc}, to end-to-end registration networks between radar
and lidar images~\cite{yin_loc}.

The reported error values are provided as root mean square errors (RMSE) in
lateral and longitudinal directions, where available. These values serve as an
indication of the order of magnitude rather than as a direct criterion for
comparison, since the map and sensor qualities as well as the test environments
differ significantly. Some work includes localization results in challenging
environments such as snow. For better comparability, these values are
not reported, but those of optimal results under regular conditions. If no
separate evaluation of lateral and longitudinal directions is available, the
joint error in the horizontal plane is given. When no RMSE values are available,
it is denoted with  ${}^\dagger$ and *. The former indicates the
\SI{68}{\percent} interval derived from a cumulative error distribution function
(equivalent to RMSE if distribution is assumed zero mean Gaussian). The symbol *
denotes the alternative error is calculated as the mean absolute error (MAE).

\subsection{Polarimetric radar in automotive applications}
\noindent
While polarimetric radar for automotive applications is experiencing increasing
interest in recent years, most of the literature deals either with the
realization of such systems~\cite{freialdenhoven_pol_sys, trummer_pol_radar} or
with improved classification of road users~\cite{visentin_pol_road_user,
  tilly_pol_road_user}. Recent publications on polarimetric synthetic
aperture radar imaging~\cite{iqbal_pol_sar, merlo_pol_sar} are closer to
localization and give a first impression of the potential benefits of
polarimetry for landmark discrimination.
Although this article's authors are not aware of any recent publication on
polarimetry-based localization, one of the early radar-based localization
implementations relies heavily on polarimetry. In order to unambiguously
identify manually placed beacons for localization, the beacons are designed to
show a unique polarimetric signature to the radar~\cite{clark_loc}. In their
subsequent work, the radar's polarimetric capability helps to recognize
characteristic natural landmarks by their even bounce scattering response
\cite{clark_even_bounce}.

The authors of this publication have published several papers related to
polarimetric localization. By demonstrating the highly anisotropic scattering
behavior of typical objects in automotive environments, the need for special
handling of polarimetric data in gridmaps to realize the full potential is
proven~\cite{weishaupt_polarimetric_anisotropy}. To deal with the changing
scattering mechanisms, an approach involving incorporation of detection-level
data into a polarimetric covariance representation per grid cell is
proposed~\cite{weishaupt_covariance}. Based on this, a landmark detector is
developed that shows increasing robustness as the amount of polarimetric
information increases~\cite{weishaupt_pcfar}. In addition to the power
comparison in conventional constant false alarm rate (CFAR) detectors, the
proposed detector includes the polarimetric scattering difference in its
decisions, which is why it is referred to as a polarimetric CFAR (pCFAR) in the
following. Finally, a gridmap construction technique taking advantage of
low-level data is introduced to represent the static environment as completely
as possible, although still non-polarimetric~\cite{weishaupt_precfar}. The
designation of low-level spectral data as PreCFAR data is introduced in this
paper emphasizing the raw nature of this data level being tapped off
before any detector stage.

This paper introduces the novel combination of low-level PreCFAR data
and polarimetric covariance gridmaps bringing together the most complete
environment representation with a sensible inclusion of polarimetric
characteristics. Furthermore, this paper is the first to analyze the impact of
different levels of polarimetric information on automotive localization
accuracy.

\begin{figure*}
  \resizebox{\textwidth}{!}{\input{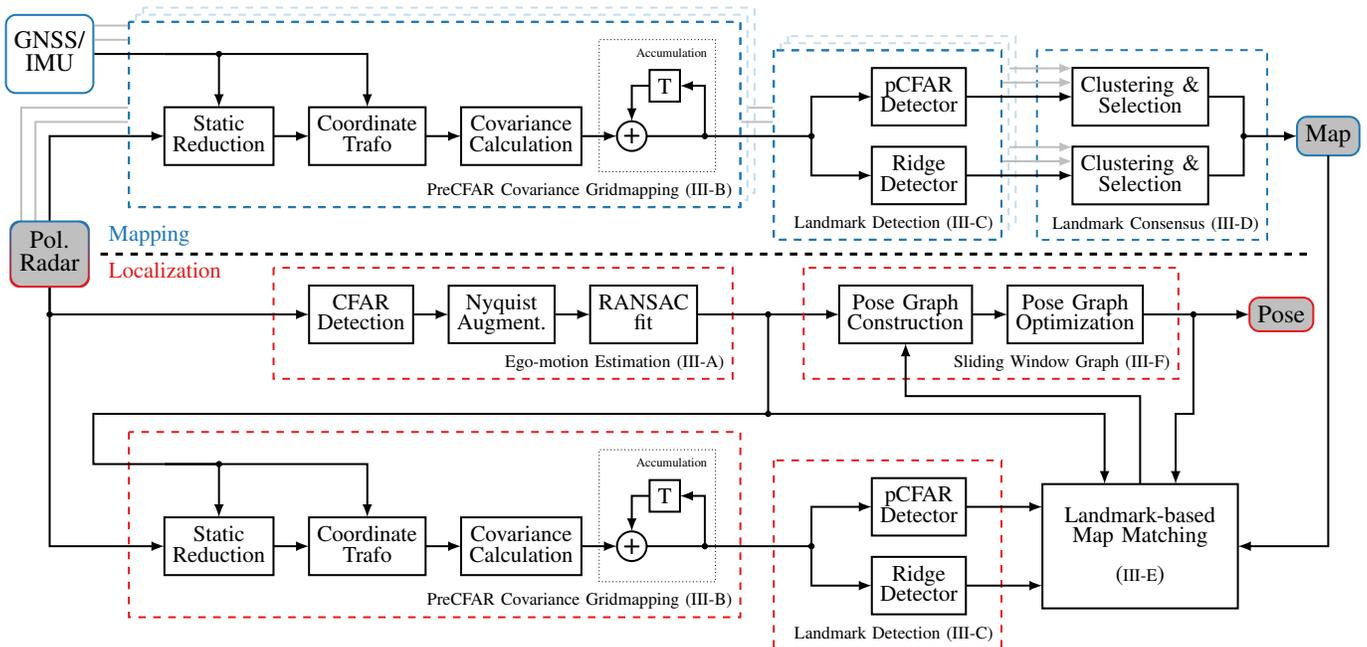}}\vspace{-0.1cm}
  \caption{The block diagram of the proposed approach contains the
    \textcolor{Paired-B}{mapping} (upper) as well as the
    \textcolor{Paired-F}{localization} part (lower). Lighter colored blocks and
    connections symbolize the multiple drives through the same environment for
    map creation, which are merged in the last mapping step.}
  \label{fig:block_diagram}
\end{figure*}

\section{PROPOSED APPROACH}
\label{sec:proposed_approach}

\noindent
In this section, the proposed self-localization approach is described in detail.
It is based on point- and line-shaped landmarks such that the corresponding maps
are compact, which is necessary for a scalable approach as mentioned above. The
main challenge to overcome for achieving a precise and reliable localization is
a robust and repeatable landmark detection. Only then can the right association
of observed landmark candidates with map landmarks be made. For this, an
accumulation of multiple frames is necessary in radar applications because
otherwise a lot of false positive point-shaped landmark candidates would be
extracted due to noise and only partial line-shaped landmarks could be
identified due to the single observation point. Furthermore, the robustness
shall be improved by leveraging polarimetric information. The proposed
implementation is developed to fulfill the combination of these requirements and
all the necessary building blocks are discussed in the following subsections.
The block diagram in Fig.~\ref{fig:block_diagram} gives an overview of the parts
involved.

\subsection{Ego-motion Estimation}
\label{sec:em_estimation}

\def\azimuthangle{\phi}
\def\dopplervelocity{v_r}
\def\egomotionstate{\mathbf{m}_\textnormal{ego}}
\def\egovelocity{v_\textnormal{ego}}
\def\egoyawrate{\omega_\textnormal{ego}}
\def\egovelocityest{v_\textnormal{ego,est}}
\def\egoyawrateest{\omega_\textnormal{ego,est}}
\def\egovelocitygt{v_\textnormal{ego,gt}}
\def\egoyawrategt{\omega_\textnormal{ego,gt}}

\noindent
The instantaneous estimation of the ego-vehicle's motion state based on Doppler
radars is well-known for automotive applications. The static environment is
observed by the radar sensors with a relative velocity that is of the same
magnitude as the vehicle's velocity, but with an opposite sign. Because the
sensors measure only the radial component of the velocity, a sinusoidal
dependence over the observation angle is expected. By estimating the sinusoid's
parameters and combining them with the mounting pose of the sensor, one can
infer the motion state. For this, reflections from the static surroundings need
to be distinguished from other dynamic traffic participants. A random sample
consensus (RANSAC) technique is suitable for this~\cite{kellner_egomotion}. It
is the basis of the ego-motion estimation in this work. %
A motion state with two degrees of freedom
$\egomotionstate = \left( \egovelocity, \egoyawrate \right)$ referenced to the
vehicle's rear-axle center including a forward velocity and yaw rate is
considered sufficient (zero-side-slip assumption). Any errors due to a violation
of this assumption are considered small enough that the subsequent localization
can handle and compensate those. Limiting accelerations to a range of physically
reasonable values allows rejecting outliers that may occur when the static
environment is not the origin of the majority of detections, e.g. in dense
traffic. Because the algorithm is sensitive to mounting pose errors, the
calibration has to be carried out carefully. Alignment between the polarimetric
channels is one of the peculiarities that have to be taken into account.
However, due to this work's focus, the reader is referred
to~\cite{weishaupt_pol_cal} for details.

Compared to the original work~\cite{kellner_egomotion}, some additional steps
help to improve accuracy and robustness in a real-world application.
For non-negligible accelerations, motion estimation including multiple sensors
can suffer from synchronization errors due to a sequential triggering for
interference avoidance. Extrapolation of motion information from previous
estimates is used to minimize this effect~\cite{juergens_localization}.
Thereby, the respective last measurements of each sensor can be included in a
combined estimation for more robust and precise results.

\begin{figure}[t]
  \resizebox{\columnwidth}{!}{\input{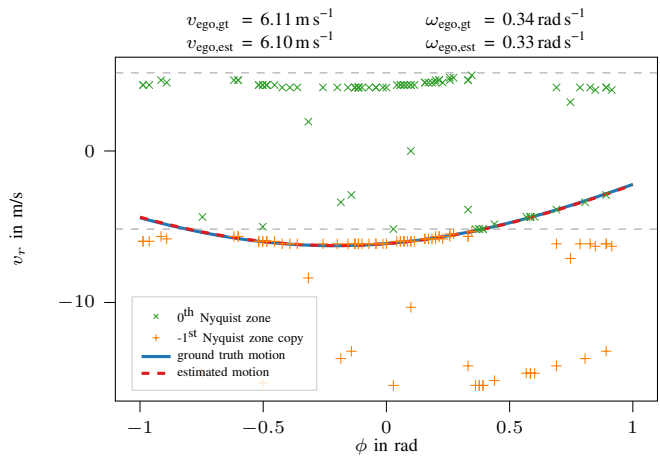}}
  \caption{By the proposed Nyquist zone augmentation, the ego-motion estimation
    also provides correct results beyond the unambiguous Doppler velocity.}
  \label{fig:em_doppler_unfolding}
\end{figure}

Furthermore, additional attention is needed for ego-velocities outside of the
unambiguous Doppler range. In order to measure the different components of the
scattering matrix, fully polarimetric implementations typically switch between
polarizations in a time multiplex scheme, which can result in small unambiguous
Doppler velocities (cf.~Table~\ref{tab:parameters}). A straightforward
implementation might involve carrying out all calculations in the \nth{0}
Nyquist zone, i.e. the domain in which all measured velocities including
potentially aliased ones are located. However, this would require a repeated
folding of the fit function and usage of cyclic distances for inlier
calculation. This is why an alternative augmentation approach is pursued in this
work, which might be more efficient even though the number of detections is
multiplied. Depending on the expected range of ego-velocities and the
unambiguous Doppler velocity, a number of Nyquist zones is determined that shall
be covered. Then, each detection is copied to the corresponding positions in all
zones such that the regular RANSAC algorithm without model function folding and
cyclic distance calculation can be applied. Due to the algorithm's objective to
identify the largest group of consistent velocities, the introduced invalid
samples are not misleading the estimation result. This approach of simple but
highly parallelizable operations is particularly suitable for modern, parallel
computing architectures. A simplified example (only one copy for clarity) is
shown in Fig.~\ref{fig:em_doppler_unfolding}, where the ego-velocity is slightly
above the unambiguous Doppler velocity for a single front-mounted sensor such
that the samples in broadside direction are aliased into the \nth{0} Nyquist
zone. Due to the proposed Nyquist zone augmentation, the RANSAC algorithm is
still able to produce a valid estimate that is very close to ground truth.

\subsection{PreCFAR Covariance Gridmaps}
\label{sec:gridmap}

\def\lhcpindex{\textnormal{L}}
\def\rhcpindex{\textnormal{R}}
\def\horizontalindex{\textnormal{H}}
\def\verticalindex{\textnormal{V}}
\def\basis{\mathcal{B}}
\def\lrbasis{{\basis_{\lhcpindex\rhcpindex}}}
\def\hvbasis{{\basis_{\horizontalindex\verticalindex}}}
\def\receiveindex{a}
\def\transmitindex{b}
\def\scatteringelement{S}
\def\scatteringmatrix{\mathbf{\scatteringelement}}
\def\scatteringvector{\mathbf{\Omega}}
\def\generalscatteringelement{\scatteringelement_{\receiveindex\transmitindex}}
\def\polcovarianceelement{C}
\def\polcovariancematrix{\mathbf{\polcovarianceelement}}
\def\poldim{q}
\def\sll{\scatteringelement_{\lhcpindex\lhcpindex}}
\def\slr{\scatteringelement_{\lhcpindex\rhcpindex}}
\def\srl{\scatteringelement_{\rhcpindex\lhcpindex}}
\def\srr{\scatteringelement_{\rhcpindex\rhcpindex}}
\def\cllll{\polcovarianceelement_{\lhcpindex\lhcpindex,\lhcpindex\lhcpindex}}
\def\clllr{\polcovarianceelement_{\lhcpindex\lhcpindex,\lhcpindex\rhcpindex}}
\def\cllrl{\polcovarianceelement_{\lhcpindex\lhcpindex,\rhcpindex\lhcpindex}}
\def\cllrr{\polcovarianceelement_{\lhcpindex\lhcpindex,\rhcpindex\rhcpindex}}
\def\clrll{\polcovarianceelement_{\lhcpindex\rhcpindex,\lhcpindex\lhcpindex}}
\def\clrlr{\polcovarianceelement_{\lhcpindex\rhcpindex,\lhcpindex\rhcpindex}}
\def\clrrl{\polcovarianceelement_{\lhcpindex\rhcpindex,\rhcpindex\lhcpindex}}
\def\clrrr{\polcovarianceelement_{\lhcpindex\rhcpindex,\rhcpindex\rhcpindex}}
\def\crlll{\polcovarianceelement_{\rhcpindex\lhcpindex,\lhcpindex\lhcpindex}}
\def\crllr{\polcovarianceelement_{\rhcpindex\lhcpindex,\lhcpindex\rhcpindex}}
\def\crlrl{\polcovarianceelement_{\rhcpindex\lhcpindex,\rhcpindex\lhcpindex}}
\def\crlrr{\polcovarianceelement_{\rhcpindex\lhcpindex,\rhcpindex\rhcpindex}}
\def\crrll{\polcovarianceelement_{\rhcpindex\rhcpindex,\lhcpindex\lhcpindex}}
\def\crrlr{\polcovarianceelement_{\rhcpindex\rhcpindex,\lhcpindex\rhcpindex}}
\def\crrrl{\polcovarianceelement_{\rhcpindex\rhcpindex,\rhcpindex\lhcpindex}}
\def\crrrr{\polcovarianceelement_{\rhcpindex\rhcpindex,\rhcpindex\rhcpindex}}
\def\scatteringvectorindex{k}
\def\covariancesamplesize{N}
\def\wishartdistance{d_{\textnormal{wishart}}}
\def\neighborhoodcovariance{\mathbf{\Sigma}}

\noindent
The proposed localization framework builds around a specifically engineered
representation of the static environment. This article's authors have addressed
some challenges involved with using polarimetric radar for self-localization in
a series of preceding work. While the core concepts and their motivation for
PreCFAR gridmaps on the one hand and the covariance representation on the other
hand are given in Sections~\ref{sec:precfar_gm} and~\ref{sec:covariance_gm},
references to the corresponding publications will help the reader to study the
details. Finally, the novel integration of both previous concepts is proposed in
Section~\ref{sec:precfar_cov_gm}.

\subsubsection{PreCFAR Gridmaps}
\label{sec:precfar_gm}
\noindent
For a rich representation of the surroundings that allows extracting
potential landmarks in a robust way, the accumulation of multiple measurement
cycles is a well-known technique. To account for the motion of the ego-vehicle,
this accumulation is typically carried out in a world-fixed coordinate system in
the form of gridmaps. In radar applications, the estimated quantity per grid
cell could be occupancy or reflectivity~\cite{werber_radar_gridmaps}.
Historically, the radar measurements for incorporation into the gridmap were in
a sparse detection list format as a result of CFAR filtering. This leads to an
incomplete mapping in case of low-reflectivity or extended objects because of a
low probability of detection and the CFAR's inherent masking property.
Therefore, a technique to counteract this is based on the use of lower-level
data that has not passed a CFAR filtering stage~\cite{weishaupt_precfar}. For
using this data level, either a sensor with a higher processing power than is
usual today needs to carry out the whole gridmap computation, or a higher
bandwidth interface has to be used that is able to transfer the increased data
rates. The latter is becoming possible soon with Ethernet becoming the common
standard interface to the radar and emerging compressed representations of
low-level data. An additional advantage of the more centralized architecture is
the option to perform low-level multi-sensor fusion.

\begin{figure*}
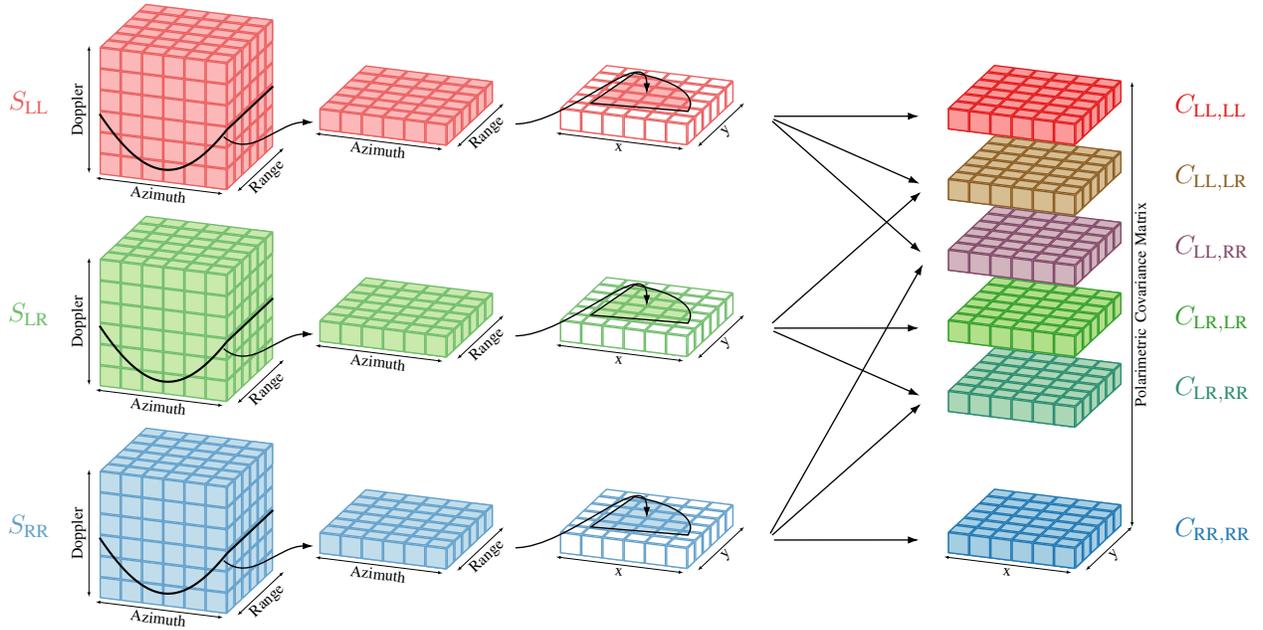

  \centering
  \include{figures/precfar_cov_gm}\vspace{-0.5cm}
  \caption{PreCFAR covariance gridmaps combine a low-level radar data
    utilization with a meaningful accumulation of polarimetric information
    for incorporation into a gridmap. As a result, the superposition of
    scattering mechanisms and extended objects are well represented.}
  \label{fig:precfar_cov_gm}
\end{figure*}

Starting from a processed radar data cube, the static surroundings are located
on a sinusoidal shaped surface as depicted in Fig.~\ref{fig:precfar_cov_gm}.
The ego-motion leads to the same dependency of the measured Doppler velocity on
the azimuth angle as described in Section~\ref{sec:em_estimation}, regardless of
the range. Applying the Fourier modulation theorem in an appropriate
way~\cite{weishaupt_precfar}, this surface is extracted from the data cube and a
dense image-like representation of the static environment in polar coordinates
is obtained (``Static reduction'' in Fig.~\ref{fig:block_diagram}). For
performing the accumulation of multiple measurements spatially correct, a
transformation to Cartesian coordinates is carried out via nearest neighbor
interpolation.
Both steps require ego-motion information. Linear and angular velocities
determine the position of the surface in the data cube, whereas a pose defines
the parameters for the coordinate transformation. In case of the mapping phase,
ground truth motion data can be used. During the localization phase, the
trajectory within a sliding window needs to be smooth and must not negatively
influence the accumulation by localization errors. For this reason, the integral
of the estimated ego-motion is used instead of the localization result
(cf.~Fig.~\ref{fig:block_diagram}, there is no feedback loop of the pose
result). The sliding window size is limited by the drift introduced by the
imperfect estimates. The maximum acceptable drift needs to be derived from the
subsequent module and should be of the same order of magnitude as the grid's
cell size in this case.

\subsubsection{Polarimetry \& Covariance Gridmaps}
\label{sec:covariance_gm}
\noindent
While conventional automotive radars measure a scalar reflectivity, usually
expressed as a radar cross section (RCS), a polarimetric radar estimates
multiple reflectivities for each cell in the data cube. Because a fully
polarized electromagnetic wave can be represented by a superposition of two
orthogonal basis polarizations, a scattering event can be fully characterized by
a 2 $\times$ 2 matrix, assuming narrow-band and far-field conditions. This can
be interpreted as two separate responses of the scattering object for each of
the two incident polarizations and consequently for any superposition due to
linearity.

Although the transmitting and receiving polarization bases may not be
the same in general, only one basis $\basis$ is assumed for both in
the following. In case of a fully polarimetric measurement, this is possible
without loss of generality because a purely mathematical transformation to any
basis is feasible. Then, the elements $\generalscatteringelement \in \mathbb{C}$
with $\receiveindex, \transmitindex \in \{\lhcpindex, \rhcpindex\}$ constitute the
complex scattering matrix $\scatteringmatrix_\lrbasis$ in a circular basis
$\lrbasis$ with a sent polarization $\transmitindex$, received polarization
$\receiveindex$ and $\lhcpindex$, $\rhcpindex$ for left- and right-hand circular
polarization, respectively. Unless otherwise noted, a circular polarization basis
$\lrbasis$ is chosen implicitly and all polarimetric quantities are given with
reference to this basis ($\scatteringmatrix = \scatteringmatrix_\lrbasis$). In
the following, the vectorized representation of the scattering matrix
\begin{equation}
  \scatteringvector = \vecfrommat\left( \scatteringmatrix \right) =
  \begin{pmatrix}
    \sll & \slr & \srl & \srr
  \end{pmatrix}^T
\end{equation}
will be used with $\scatteringvector \in \mathbb{C}^\poldim$. The dimensionality
$\poldim$ indicates if fully polarimetric data is used ($\poldim \in \{3, 4\}$),
a dual polarimetric configuration is evaluated ($\poldim = 2$) or a non/single
polarimetric experiment is carried out ($\poldim = 1$). A scattering vector of
length $\poldim = 3$ still has full polarimetric information in case of a
mono-static radar setup and reciprocal propagation media because both originally
anti-diagonal terms are identical ($\slr = \srl$). In the non-polarimetric case,
the only remaining absolute phase provides no usable information and could be
neglected ($\poldim = 1 \rightarrow \scatteringvector \in \mathbb{R}$). Due to
the used covariance formulation that is introduced below, this happens
inherently by the covariance definition. In a circular basis, $\sll$ and $\srr$
correspond to even-bounce scattering, $\slr$ and $\srl$ to odd-bounce
scattering.

While the scattering vector can properly describe the observation of a single
scattering event, it is an inappropriate representation for a superposition
of multiple scattering observations in a single gridmap cell. Such a
superposition can occur for various reasons: on the one hand, multiple objects
with different polarimetric signatures could be located in the same grid cell;
on the other, an anisotropic object that is observed from diverse aspect angles
either by multiple radars with different mounting positions or over time due to
the ego-vehicle's motion~\cite{weishaupt_polarimetric_anisotropy}. A more
suitable representation is found by modeling the scattering information as a
stochastic process via its second order moment~\cite{weishaupt_covariance}.
Therefore, a sample covariance matrix $\polcovariancematrix$ over a sample size
of $\covariancesamplesize \in \mathbb{N}$ is calculated per grid cell by
averaging the outer products of the corresponding scattering vectors
$\scatteringvector_\scatteringvectorindex$ over time and different sensors as
\begin{equation}
  \polcovariancematrix = \frac{1}{\covariancesamplesize} \sum_{\scatteringvectorindex=1}^N \scatteringvector_\scatteringvectorindex \, {\scatteringvector_\scatteringvectorindex}^H
  \label{eq:covariance}
\end{equation}
with $\polcovariancematrix \in \mathbb{C}^{\poldim\times\poldim}$ and
superscript $^H$ as the Hermitian transpose.

\subsubsection{PreCFAR Covariance Gridmaps}
\label{sec:precfar_cov_gm}
\noindent
The combination of both previously introduced concepts is novel and builds the
basis for the localization. As depicted in Fig.~\ref{fig:precfar_cov_gm}, the
different polarization channels are processed separately with the proposed
PreCFAR gridmapping technique first. At this point, one could comprehend the
deliberate choice of a simple nearest neighbor interpolation: it prevents the
introduction of distorted scattering responses that were never measured, but are
purely a result of interpolation. For reasons of efficiency, the covariance
calculation is performed {\itshape after} Cartesian PreCFAR gridmaps are created
per polarization channel. Due to the Hermitian property of the covariance
matrix, it is sufficient to compute and store only the diagonal and the upper
off-diagonal elements. Instead of $\poldim^2$ layers, the memory requirement is
reduced to $\frac{\poldim \left(\poldim + 1 \right)}{2}$ layers for the
compressed covariance format.

This processing block is used in both the mapping as well as in the localization
phase (cf.~Fig.~\ref{fig:block_diagram}). The only difference between both
instances is the persistence of old measurement data. While during mapping it is
assumed that the ground truth pose does not suffer from any drift, this is not
valid for the integrated ego-motion estimation. Therefore, for mapping, the full
history of each cell is used, which includes visits to the same area with
different driving directions. For localization, on the other hand, the finite
sliding window described in Section~\ref{sec:precfar_gm} is required to prevent
smearing.

\subsection{Landmark extraction}
\label{sec:lm_extraction}
\noindent
The landmark-based localization relies on point- and line-shaped landmarks.
Similar to the PreCFAR covariance gridmapping blocks, the same landmark
detectors are used for both mapping as well as localization. Their working
principles are detailed in the following.

\begin{figure}
  \centering
  \resizebox{\columnwidth}{!}{\input{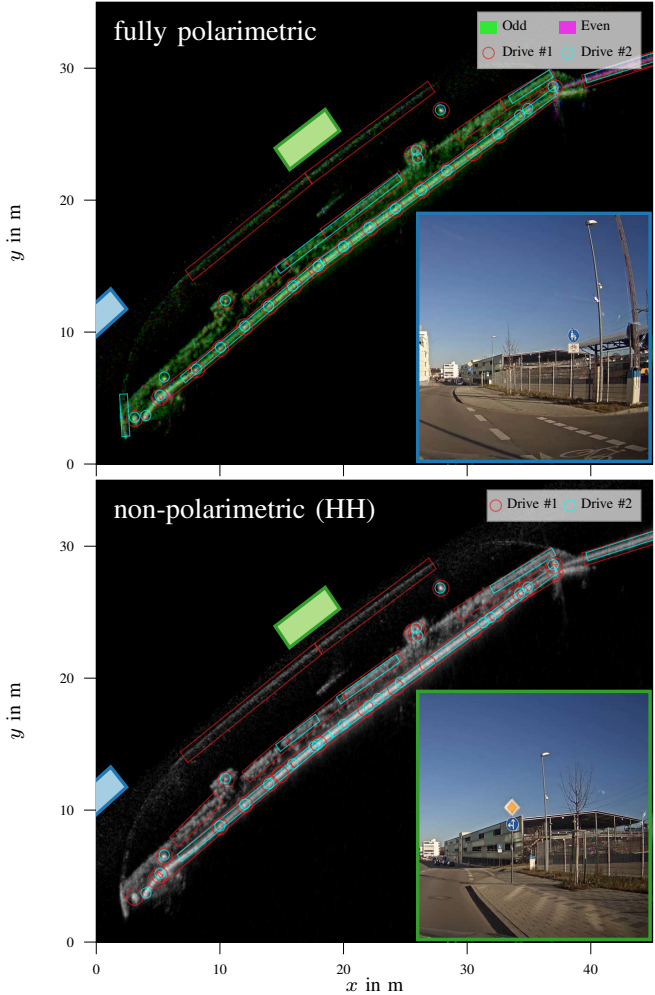}}
  \caption{The point- and line-shaped landmark candidates of the proposed
    extractors are shown for two drives in the same environment with an
    underlying PreCFAR gridmap of one of the drives. Comparing the
    non-polarimetric with the fully polarimetric experiment, the improved
    robustness of the point-shaped landmarks can be observed for the latter. The
    blue and green rectangles show the ego-vehicle positions where the optical
    images were taken.}
  \label{fig:landmark_extraction}
\end{figure}

\subsubsection{Point-shaped Landmarks}
\noindent
For the detection of point-shaped landmarks, the polarimetric information of the
covariance gridmap is exploited. This is motivated by the fact that an
additional distinct scattering information helps detecting a landmark more
robustly compared to the classical criterion of a characteristic amplitude.
Evidence for this is demonstrated in~\cite{weishaupt_pcfar} based on CFAR
gridmaps of real-world measurements. The more polarimetric information was
available to the detector, the better was the robustness of the extracted
landmarks in terms of re-observability in multiple drives. This principle is
transferred and newly applied to PreCFAR gridmaps in this paper and an example
is given by Fig.~\ref{fig:landmark_extraction}. Using the amplitude as the sole
quantity to extract characteristic areas in the non-polarimetric case, only a
few fence posts are detected. Additionally, the result is unstable as the
intersection of locally corresponding landmarks is small for multiple visits of
the same area. As the reflectivities of the fence elements and the posts are
roughly similar, the extraction algorithm has no chance to generate robust
results. The opposite is true if the characterization of the scattering
mechanism by polarimetry is included. On a closer look, the posts show
additional even-bounce scattering behavior such that a suitable algorithm is
capable to robustly identify them as landmarks.

The underlying principle is based on a rigorous continuation of a probabilistic
modeling of the polarimetric scattering vector. Starting from a multivariate
complex Gaussian distribution with zero mean for the scattering vector
$\scatteringvector$, a complex Wishart distribution can be derived as the
probability density of the covariance matrix
$\polcovariancematrix$~\cite{lee_wishart_distance}. Combining this with Bayesian
maximum likelihood classification theory, a simple distance measure

\begin{equation}
  \wishartdistance = \poldim^{-1} \left( \ln |\neighborhoodcovariance|+\operatorname{tr}\left(\neighborhoodcovariance^{-1} \polcovariancematrix \right) \right)
  \label{eq:wishart_distance}
\end{equation}
allows one to compare the similarity of polarimetric covariance matrices with
$|\mathbf{X}|$ and $\operatorname{tr}\left( \mathbf{X} \right)$ symbolizing the
determinant and trace of matrix $\mathbf{X}$,
respectively~\cite{weishaupt_pcfar}. For landmark detection, this is applied in
a similar fashion to cell averaging CFAR by comparing the covariance of the cell
under test $\polcovariancematrix$ with a neighborhood average
$\neighborhoodcovariance$, referred to as pCFAR. A threshold binarizes the
resulting distance image and the centroids of connected areas are considered
potential landmarks. For details, the reader is referred to
\cite{weishaupt_pcfar}.

\subsubsection{Line-shaped Landmarks}
\label{sec:line_lm_extraction}
For line-shaped landmark extraction, no polarimetric information is used because
environments in which a significant benefit can be expected typically do not
occur (extended object within another extended object similar in amplitude but
different in scattering). Instead, the trace of the covariance matrix as a
scalar measure of reflection power is the input to the algorithm. The first
operation on this map is median filtering to reduce noise while preserving
structure. Then, ridge detection  is carried out using the eigenvalues of the
Hessian matrix. The computed ridgeness score is thresholded and the result is
skeletonized. Finally, a Hough transform extracts line segment candidates.
Comparing the line-shaped landmark candidates in
Fig.~\ref{fig:landmark_extraction}, no general difference is observed between
the different levels of polarimetric information utilization. This is partly
expected as no additional scattering information is used. Furthermore, it shows
that the lowered power in the non-polarimetric case constructed by omitting some
channels is not the limiting factor in this scenario.

\subsection{Landmark Consensus \& Map Construction}
\label{sec:landmark_consensus}
\noindent
For map construction, the area of interest is passed multiple times with a
known ground truth pose from a coupled system consisting of an inertial
measurement unit (IMU) and a differential GNSS receiver. Hence, map
construction in this work only considers scenarios in which the outage of
satellite signal reception is limited to short time periods. Ground truth
poses are also necessary to evaluate the localization accuracy. However, that
does not compromise the proposed localization approach itself, which would be
most valuable in GNSS denied scenarios. It would just require other mapping
methods mapping, e.g. incorporating SLAM or alternative ground truth positioning
solutions, which is beyond the scope of this paper.

The stacked copies of the processing blocks in the lighter blue color in the
mapping path of Fig.~\ref{fig:block_diagram} symbolize the multiple drives for
which a consensus has to be found. The rationale behind finding landmarks as an
agreement over multiple drives is motivated by the assumption that a landmark is
robust in the sense of easy recognition if it is seen multiple times. At the
same time, this includes the removal of only temporarily stationary objects,
such as parked cars or trash bins, which are not suitable as map landmarks.

For point-shaped landmarks, this is achieved by evaluation of the Euclidean
distances between landmark candidate positions of the different passes. The
landmark candidate positions of each combination of passes are queried for
spatially close pairs via an efficient $k$-d tree data
structure~\cite{bentley_kd_tree}. By interpreting the landmarks as nodes and the
aforementioned closeness below a threshold $d_{\pointlandmarknode,
  \textnormal{thres}}$ as edges in a graph, a connected component algorithm is
used to establish a landmark from a group of landmark candidates if a minimum
number of observations $N_{\pointlandmarknode, \textnormal{thres}}$ is exceeded.
The final landmark position is simply the average position of all group members.

Finding line-shaped landmarks is conceptually similar to the connected component
merging of re-observed candidates with the difference of the distance measure.
Instead of the Euclidean distance a combination of convex hull and projection
distance is deployed and proximity is assumed when the corresponding values fall
below the thresholds $d_{\linelandmarknode,\textnormal{ch,thres}}$ and
$d_{\linelandmarknode,\textnormal{prj,thres}}$~\cite{werber_line_distance}.
Identified matching groups of candidates are merged as a line segment with the
longest distance between all mutual projection points
(cf.~\cite{werber_line_distance}). In case a series of line segments describing
a curve form a group, shortcuts have to be prevented. Therefore,
additional points are inserted in a similar fashion to the Ramer-Douglas-Peucker
algorithm~\cite{ramer_algo} until an upper error threshold is met. Lastly, if
the line segments' ends cross or are just short of crossing, they are shortened
or lengthened to the intersection point for a more compact representation in the
map.

\subsection{Landmark-based Map Matching}
\label{sec:map_matching}
\noindent
The robustness of the whole localization framework depends strongly on the
map matching module. It has to unambiguously associate the potential landmark
observations with actual landmarks in the map. The challenges to be overcome
include finding the correct associations even in the case of partial overlap
between both sets, and avoiding getting stuck in a local minimum, as observed in
some common registration algorithms.
Because this task is to some extent independent of the sensor modality used,
there is a wider range of literature available. This allowed the adoption of
many ideas from the initially modality-independent work~\cite{wilbers_diss} on a
robust map-based self-localization framework. The evaluation of the framework is
based on camera and lidar data in this publication, but an application of this
to radar is also published~\cite{juergens_localization}.

First, landmark observations are collected in a local map frame that is purely
based on the integration of ego-motion estimates (i.e. not result of the full
localization to avoid self-reinforcing divergence). Therein, observations within
a temporal window belonging to the same landmark are grouped according to their
mutual distances. In contrast to~\cite{wilbers_diss}, the clusters are
calculated by finding connected components in a distance-weighted sparse graph
derived from a $k$-d tree representation. This allows larger time windows for
local map construction, because more drift is tolerable due to the connected
component clustering. In general, the local map clustering enables correct
association in case of a temporarily low number of landmarks observations.

Next, the local landmark clusters obtained in this way are matched with the
map landmarks analogously to~\cite{wilbers_diss}. Both, point- and line-shaped
landmarks contribute to the cost of a transformation candidate. As introduced in
Section~\ref{sec:landmark_consensus}, the Euclidean distance specifies the cost
of point-shaped landmarks, whereas the sum of convex hull and projection
distance is used for line-shaped landmarks. To avoid low cost but also low count
matches, an additional cost is introduced for unmatched landmarks.

Finally, the temporal smoothing is again conceptually very similar
to~\cite{wilbers_diss} by generally searching for the map landmark that was most
often associated with a landmark observation. This filters out outlier matches.
Note that this allows a map landmark to be associated with multiple
observations. Therefore, the proposed implementation differs by aiming at an
injective mapping between the observations and map landmarks. If there is a
landmark observation without a correct counterpart in the map, and it is wrongly
associated a few times to some map element, which actually corresponds to
another landmark observation with a higher number of associations, both
associations would outlast the temporal smoothing in the original formulation.
Instead, the desired result is a maximum weight matching of a bipartite graph
constructed from the corresponding association counts. %
Then, no map landmark will be associated with multiple observations after the
temporal smoothing operation.

\begin{figure}
  \centering
  \resizebox{\columnwidth}{!}{\input{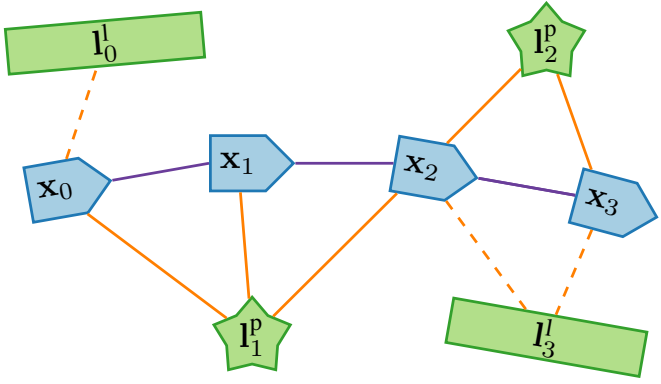}}
  \caption{The result of the map matching module is an association graph that
    connects point- (\textcolor{Paired-H}{\solidrule}) and line-shaped
    (\textcolor{Paired-H}{\protect\dashedrule}) landmark candidate observations
    with their corresponding map landmarks. Based on this, the pose graph is
    constructed for the subsequent optimization step
    (cf.~Fig.~\ref{fig:pose_graph}).}
  \label{fig:association_graph}
\end{figure}

Based on the results of the described map matching approach, an association
graph can be drawn for visualization. Poses $\posenode_\poseindex \in
\liegrouprigid{2}$ ($\poseindex \in \{0, \ldots, \numberposes - 1\}$)
are connected by the integrated odometry in their temporal order. If the
result of the map matching is that a landmark $\landmarknode_\landmarkindex$
($\landmarkindex \in \{0, \ldots, \numberlandmarks - 1\}$) is associated with an
observation in some measurement cycle, the corresponding pose is connected to
the landmark. Point-shaped landmarks are denoted with
$\pointlandmarknode_\landmarkindex \in \mathbb{R}^2$ and line-shaped landmarks
with $\linelandmarknode_\landmarkindex \in \mathbb{R}^4$. A simple example of
such an association graph is shown in Fig.~\ref{fig:association_graph}.

\subsection{Sliding Window Pose Graph Optimization}
\label{sec:sliding_window_pose_graph_optimization}
\noindent
The pose estimate of the proposed approach is derived within a factor graph
optimization framework. Again, this subsection is inspired by publication
\cite{wilbers_diss}. However, the graph formulation is an established technique
and the cited reference is one of many. Furthermore, the handling of line-shaped
landmarks is different from that in the cited publication and therefore the
proposed approach is discussed in more detail below.

\begin{figure}
  \resizebox{\columnwidth}{!}{\input{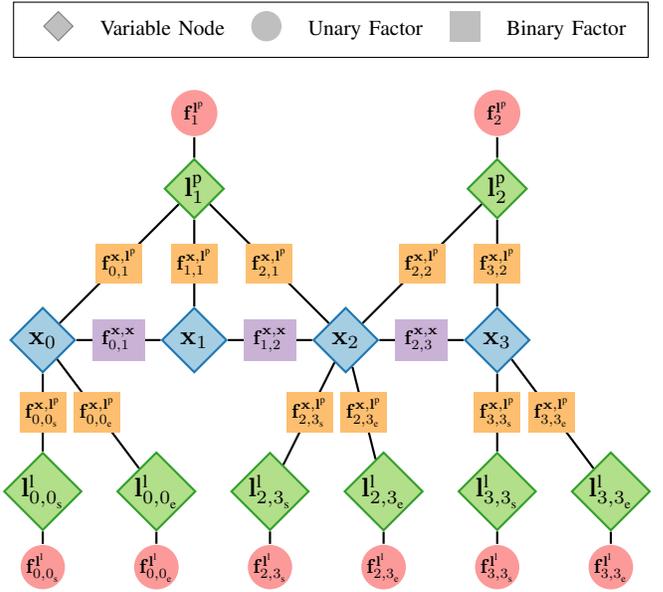}}
  \caption{The pose graph consists of nodes whose values are to be optimized,
    and factors that constrain the optimization. The ego-vehicle poses
    $\posenode$ are the main nodes to be optimized whereas optimized landmark
    nodes $\landmarknode$ are of more interest for map updates and less for
    the localization itself. This example is the equivalent pose graph to the
    association graph of Fig.~\ref{fig:association_graph}.}
  \label{fig:pose_graph}
\end{figure}

Several properties make graph-based approaches advantageous for automotive
self-localization: they have deterministic behavior and are able to integrate
time-delayed information after ambiguity resolution. %
Although the intended application is localization, the landmark positions are
still part of the estimation result in addition to the poses. This allows
updating the map, but is beyond the scope of this work. The size of the factor
graph is kept fixed at a certain duration in form of a sliding window. Due to
the localization focus, outdated nodes are removed from the graph by truncation
instead of marginalization to avoid fill-in \cite{wilbers_diss}. %
According to this, the state to optimize $\graphvariables = \begin{pmatrix}
\posenode^T & \landmarknode^T \end{pmatrix}^T$ consists of $\numberposes$~poses
$\posenode = \begin{pmatrix}\posenode_0^T & \cdots & \posenode_{\numberposes-1}^T
\end{pmatrix}^T$ and %
$\numberlandmarks$~landmarks $\landmarknode = \begin{pmatrix}\landmarknode_0^T
& \cdots & \landmarknode_{\numberlandmarks-1}^T\end{pmatrix}^T$. These
correspond to the variable nodes in the factor graph representation
(cf.~Fig.~\ref{fig:pose_graph}). The objective of the optimization is to find
the state $\hat{\graphvariables}$ that is most probable given the measurements
$\measurement$ in a maximum a posteriori sense as
\begin{equation}
  \hat{\graphvariables} = \argmax_{\graphvariables} \; p\left( \graphvariables \mid \measurement \right).
\end{equation}

Applying Bayes theorem and assuming independent and identically distributed
Gaussian densities for the measurements and priors, the optimization can be
reformulated in form of a minimization as
\begin{equation}
  \hat{\graphvariables} = \argmin_{\graphvariables} \sum_k \errorfunction_k\left( \graphvariables, \measurement_k \right)^T \, \informationmatrix_k \,\errorfunction_k\left( \graphvariables, \measurement_k \right)
\end{equation}
with the error function $\errorfunction_k$ and information matrix
$\informationmatrix_k$ as the inverse of the covariance matrix corresponding to
measurement $\measurement_k$. Factor nodes $\graphfactor_k$ in a factor graph
comprise the error function $\errorfunction_k$, information matrix
$\informationmatrix_k$ as well as the measurement $\measurement_k$ and are the
other part in the factor graph representation (cf.~Fig.~\ref{fig:pose_graph}).
While the index $k$ in this summation stands for any existing factor to be
considered, the nomenclature for a specific factor is as follows. If a factor is
directly influencing only one variable, this {\itshape unary} factor
$\graphfactor^u_i$ is in relation with the variable node of type $u \in
\{\posenode, \pointlandmarknode, \linelandmarknode \}$ at index $i$.
Analogously, for a {\itshape binary} factor $\graphfactor^{u,v}_{i,j}$ an
additional variable node of type $v \in \{\posenode, \pointlandmarknode,
\linelandmarknode \}$ at index $j$ is directly influenced. One can imagine them
as containing a relative measurement between the involved nodes. Thus, four
factor types are relevant for the localization task being

\begin{itemize}
  \item $\graphfactorposepose$: binary odometry factor,
  \item $\graphfactorposepointlandmark$: binary landmark observation factor,
  \item $\graphfactorpointlandmark$: unary point-shaped map landmark prior factor,
  \item $\graphfactorlinelandmark$: unary line-shaped map landmark prior factor.
\end{itemize}

Translation from the association result in Fig.~\ref{fig:association_graph} to a
factor graph representation is shown in Fig.~\ref{fig:pose_graph}. The
radar-based ego-motion estimate constrains successive pose variables.
Forward and rotational velocities are integrated by an ordinary linear model to
a position and heading delta, which serves as the factor's measurement
$\measurementposepose$. For simplicity, the corresponding information matrices
$\informationmatrixposepose$ are chosen to be constant, and the error function
is typically implemented in any graph optimization library via Lie group
calculus.

Landmark observation measurements $\measurementposepointlandmark$ and their
corresponding information matrices $\informationmatrixposepointlandmark$
are often considered in polar coordinates to account for the sensor's
uncertainties in the coordinate system that relates to its measurement
principle. Again, the information matrices for all factors are constant in the
polar coordinate system, and derivation is based on the sensor's accuracy and
precision. Aspect angle dependencies are neglected. These binary factors are
used for both point- as well as line-shaped landmarks with the latter being
split into start and end point as described in the following.

Associated point-shaped map landmark positions constrain the landmark
state with the global map position, which is introduced in form of a measurement
$\measurementpointlandmark$ within an unary factor $\graphfactorpointlandmark$.
For these factors, the Euclidean distance serves as the error function and a
constant information matrix $\informationmatrixpointlandmark$ over all
point-shaped map factors represents the map's uncertainty.

\begin{figure}
  \resizebox{\columnwidth}{!}{\input{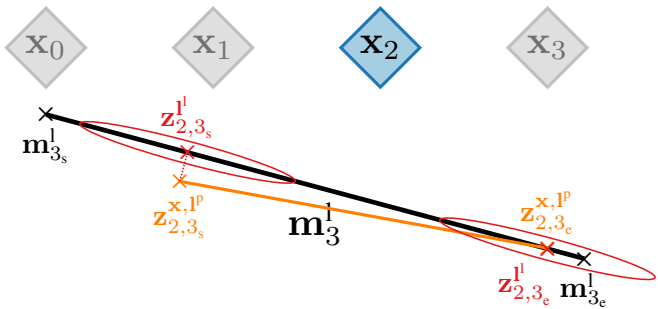}}
  \caption{The implementation of line-shaped landmark constraints is done by
    introducing two point-shaped landmark factors at the orthogonal
    projection points of the observation's start and end point onto the map
    landmark. The corresponding information matrices ensure that line-shaped
    landmark factors constrain mainly in a direction orthogonal to the map
    landmark's orientation.}
  \label{fig:line_landmark_projection}
\end{figure}

Line-shaped landmarks $\linelandmarknode = \begin{pmatrix}
\left( \linelandmarknode_\startnodesubscript \right)^T & \left( \linelandmarknode_\endnodesubscript \right)^T
\end{pmatrix}^T$ with $\linelandmarknode_\startnodesubscript,
\linelandmarknode_\endnodesubscript \in \mathbb{R}^2 $ and with the subscripts
$\startnodesubscript$ and $\endnodesubscript$ denoting the start and end point
of the line segment are handled as two point-shaped landmarks with tailored
information matrices.

For the line-shaped landmark $\linelandmarknode_3$ in
Fig.~\ref{fig:association_graph}, the observations per time instance are split
into two point observations each in the pose graph in Fig.~\ref{fig:pose_graph}.
Extra nodes for observation from pose $\posenode_2$ and pose $\posenode_3$ are
inserted because different segments of the line landmark may be observed at
different instances of time, e.g. due to occlusion.
Fig.~\ref{fig:line_landmark_projection} illustrates how the map constraints are
introduced for observation from pose $\posenode_2$. The start and end point of
the observation are handled by two independent factors that are equivalent to
the point-shaped landmark factors. Hence, the measurements are
$\measurementposepointlandmark_{{2,3}_\startnodesubscript}$ and
$\measurementposepointlandmark_{{2,3}_\endnodesubscript}$. Because the previous
association step will only deliver a positive result when the match is rather
good, it is reasonable to introduce the map constraints as the orthogonal
projections onto the line landmark. The particularity is that the information
matrices are chosen in a way that respects the line segments nature by strongly
penalizing errors in the orthogonal direction compared to parallel
(cf.~Fig.~\ref{fig:line_landmark_projection}). To indicate this, the map
constraint factors $\graphfactorlinelandmark$ are having the line landmark
superscript even though the variable nodes of start and end point have point
landmark dimensionality. In order to distinguish between the start and end point
of a line-shaped map landmark and the constraints for a specific observation,
the former are denoted by $\maplinelandmarknode$ as opposed to the notation of
measurements $\measurement$ corresponding to the factors in
Fig.~\ref{fig:line_landmark_projection}.

\section{EXPERIMENTAL EVALUATION}
\label{sec:experimental_evaluation}

\subsection{Experimental Vehicle Setup}
\label{sec:measurement_setup}
\begin{figure}
  \centering
  \input{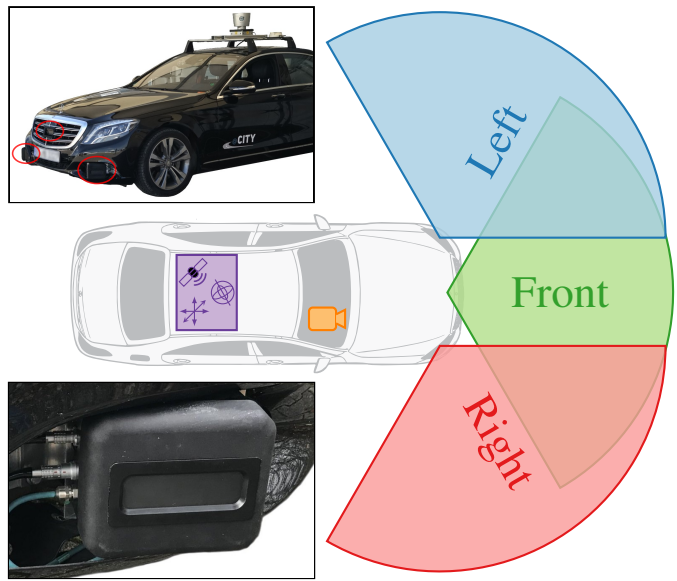}\vspace{-0.4cm}
  \caption{The experimental vehicle is equipped with three polarimetric radar
    prototypes mounted in the front, a camera for documentation purposes and an
    integrated GNSS/IMU device for map creation and localization evaluation.
    No sensor other than radar is required for a localization pass, if no
    evaluation comparing to a ground truth is carried out, which is why the
    proposed approach is considered to be radar-only during localization. (upper
    left image: original photograph is property of @CITY consortium)}
  \label{fig:measurement_setup}
\end{figure}

\noindent
The three polarimetric radar sensors are mounted in the experimental vehicle's
front. An illustration of the measurement setup as well as photographs of the
vehicle and sensor are depicted in Fig.~\ref{fig:measurement_setup}. The radar
parameters are listed in Table~\ref{tab:parameters}. The dual circularly
polarized transmit and receive antennas are implemented as corrugated horn
antennas with septum polarizers in a waveguide front-end. Fully polarimetric
information is acquired by concurrently receiving left and right hand circular
polarization and switching the transmitter between left and right hand
polarization sequentially as in time division multiplexing (TDM). This requires
a compensation of the phases of the two halves of the scattering matrix.
Otherwise, the perceived polarimetric information of the same scattering
mechanism would depend on the relative velocity between the radar and the
object. A compensation approach has been validated in~\cite{weishaupt_pol_cal}.
Additional multiplexing in time domain is employed to extend the virtual array
by a MIMO processing scheme. As a result, the unambiguous Doppler velocity is
comparatively low. After demultiplexing, all four polarimetric channels are
available at each virtual array position%
~\cite{weishaupt_pol_cal}. The raw analog-to-digital converter samples are
available per Ethernet and allow processing the proposed PreCFAR gridmaps.

Furthermore, a camera is included for documentation purposes. For map
construction and localization evaluation only, a coupled GNSS/IMU system is
installed (Genesys Automotive Dynamic Motion Analyzer G PRO). Its data is not
used in the localization pass itself. Post-processing improves the accuracy of
the pose information by forward-backward smoothing. For evaluation of the
localization, it is differentiated between a full-precision differential carrier
phase solution and lower quality solutions. The device reports expected position
errors in form of standard deviations from its internal Kalman filter. Based on
this, the mean position standard deviation in the horizontal plane over all
evaluated measurements is \SI{14}{\milli\meter} for carrier phase solutions and
\SI{86}{\milli\meter} for the rest.

\begin{table}[b]
  \renewcommand{\arraystretch}{1.4}
  \caption{Parameters of the used polarimetric radar sensors. Resolution values
  are based on the Rayleigh criterion.}
  \label{tab:parameters}
  \centering
  \resizebox{\columnwidth}{!}{%
    \begin{tabular}{lc||lc}
      \toprule
      \bfseries Parameter & \bfseries Value & \bfseries Parameter & \bfseries Value\\
      \midrule
      Center Frequency & \SI{78}{\giga\hertz} & Azimuthal Field of View & \SI{\pm 60}{\degree}\\
      Maximum Range & \SI{80}{\metre} & Azimuthal Resolution & \SI{\sim 3}{\degree}\\
      Bandwidth & \SI{2}{\giga\hertz} & Unique Virtual Elements & 31 \\
      Range Resolution & \SI{7.5}{\centi\metre} & Unambiguous Rad. Vel. & \SI{\pm 5}{\metre\per\second}\\
      Update Rate & \SI{10}{\hertz} & Rad. Vel. Resolution  & \SI{0.15}{\metre\per\second}\\
      Polarimetric MIMO & TDM & Polarization Basis & $\lhcpindex\rhcpindex$\\
      \bottomrule
    \end{tabular}}
\end{table}

\subsection{Evaluation Implementation}
\noindent
The localization performance is evaluated after the selected environment has
been traversed multiple times. A separate map is constructed for each
evaluation pass that includes all drives except its own one in the
landmark finding process. This avoids an unfair evaluation. The following
parameters are empirically derived and used in this process:
$d_{\pointlandmarknode, \textnormal{thres}} = \SI{0.3}{\meter}$,
$N_{\pointlandmarknode, \textnormal{thres}} = 2$,
$d_{\linelandmarknode,\textnormal{ch,thres}} = \SI{0.09}{\meter}$,
$d_{\linelandmarknode,\textnormal{prj,thres}} = \SI{1.5}{\meter}$.
For graph optimization, the open source library \mbox{GTSAM}~\cite{gtsam} is
used. All required factor types have implementations of the error functions
shipped with the library, so the following are used:
\texttt{BetweenFactorPose2}, \texttt{BearingRangeFactor2D} and
\texttt{PriorFactorPoint2}. A \texttt{PriorFactorPose2} is introduced in case
the number of observed and associated landmarks is insufficient and the global
vehicle pose is underconstrained. The prior factor constrains the last pose node
within the sliding window with a previously derived pose estimate such that the
current estimate is effectively the integration of the odometry factors. For the
information matrices of the factors, empirically found uncorrelated standard
deviations were chosen. The values can be found in
Table~\ref{tab:information_matrices}.
\begin{table}
  \renewcommand{\arraystretch}{1.4}
  \caption{Information Matrix Parameters}
  \label{tab:information_matrices}
  \centering
  \resizebox{\columnwidth}{!}{%
    \begin{tabular}{ll}
      \toprule
      \bfseries Information Matrix & \bfseries Parameter \\
      \midrule
      $ \informationmatrixpointlandmark = \operatorname{diag} \left({\sigmapointlandmark}^2, {\sigmapointlandmark}^2 \right)^{-1} $ & \multilinecell{$\sigmapointlandmark = \SI{0.1}{\meter}$}\\
      \hline
      $ \informationmatrixlinelandmark = {\left( \mathbf{U} \operatorname{diag} \left({\left( 10 \sigmalinelandmark \right)}^2, {\sigmalinelandmark}^2 \right) \mathbf{U}^T \right)}^{-1} $ & \multilinecell{$\sigmalinelandmark = \SI{0.1}{\meter}$\\$\mathbf{U}$ rotation acc. to Fig.~\ref{fig:line_landmark_projection}}\\
      \hline
      ${\informationmatrixposepose} = \operatorname{diag} \left({\sigmaposeposetranslation}^2, {\sigmaposeposetranslation}^2, {\sigmaposeposerotation}^2 \right)^{-1} $ & \multilinecell{$\sigmaposeposetranslation =\SI{0.05}{\meter}$\\$\sigmaposeposerotation = \SI{0.01}{\radian}$}\\
      \hline
      ${\informationmatrixposepointlandmark} = \operatorname{diag} \left( {\sigmaposepointlandmarkbearing}^2, {\sigmaposepointlandmarkrange}^2 \right)^{-1} $ & \multilinecell{$\sigmaposepointlandmarkrange = \SI{0.5}{\meter}$\\$\sigmaposepointlandmarkbearing = \SI{0.03}{\radian}$}\\
      \bottomrule
    \end{tabular}}
\end{table}

Because each iteration of the localization loop is an incremental update of the
pose graph, the iSAM2~\cite{kaess_isam} implementation is used, which is
efficient for incremental graph optimization by taking advantage of a Bayes tree
data structure. In each measurement cycle, the point-shaped landmark
observations are queried for map counterparts and found matches are incorporated
into the graph. Contrary, line-shaped landmarks are considered only in every
\nth{10} frame or when the number of point-shaped matches is less than~5. While
this has no negative impact on the localization performance according to the
experimental validation, it saves computational time because the line-shaped
landmark distance measure calculation is more demanding compared to point-shaped
landmarks and an increase of the graph's node size occurs for each observation
(cf. Section~\ref{sec:sliding_window_pose_graph_optimization}). While this
work's focus is not on a highly optimized implementation of the proposed
approach, attention is paid to only use algorithms that are able to run in a
vehicle from a complexity point of view. The recent trend of increasingly
widespread graphics processing units for highly parallel operation in the
automotive context is taken into account. All radar signal processing and
gridmapping is implemented in CUDA and real-time capable. However, the remaining
parts of the building blocks in Fig.~\ref{fig:block_diagram} are implemented in
Python and the evaluation is run offline.

\subsection{Dataset}

\begin{table}
  \renewcommand{\arraystretch}{1.2}
  \caption{Overview of routes used for evaluation}
  \label{tab:route_overview}
  \begin{tabularx}{\columnwidth}{lccl}
    \toprule
    {\bfseries Route} & {\bfseries \makecell[c]{Single-Pass\\Route Length}} &
    {\bfseries \makecell[c]{Evaluated Ride\\Duration}} & {\bfseries \makecell[c]{Environment\\Type}}\\
    \midrule
    Frankfurt & \makebox[\widthof{\SI{10}{\kilo\meter}}][r]{\SI{7}{\kilo\meter}} & \SI{45}{\minute} & urban, industrial\\
    Ingolstadt & \SI{10}{\kilo\meter} & \SI{75}{\minute} & urban, industrial\\ %
    Ulm & \makebox[\widthof{\SI{10}{\kilo\meter}}][r]{\SI{5}{\kilo\meter}} & \SI{35}{\minute} & village, rural \\
    \bottomrule
  \end{tabularx}
\end{table}

\begin{figure*}
  \pgfkeys{/pgf/number format/.cd,1000 sep={}}
  \centering
  \resizebox{\textwidth}{!}{\input{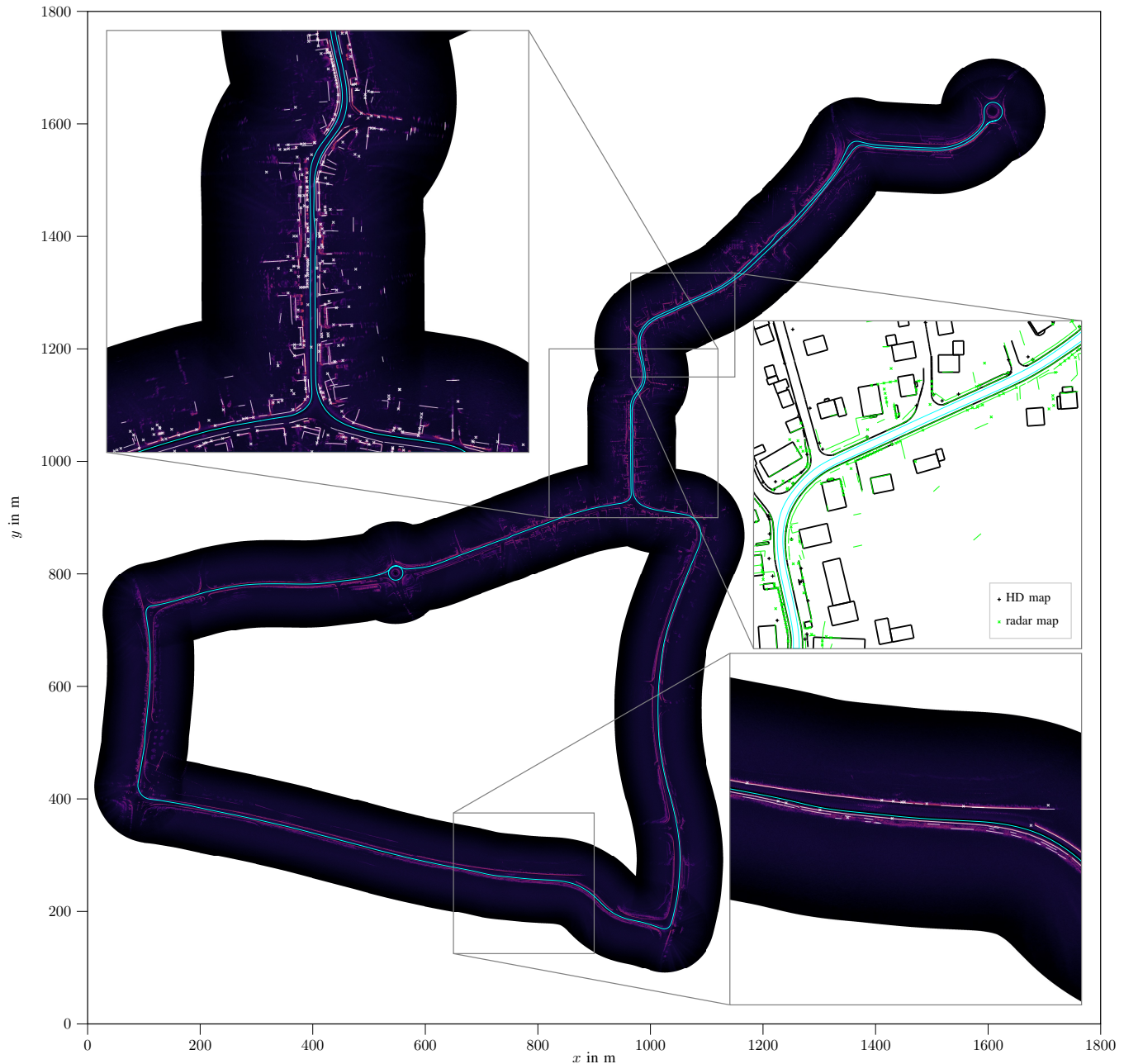}}\vspace{-0.1cm}
  \caption{The Ulm route is shown as an amplitude PreCFAR gridmap. Due to
    the high resolution, the reader is invited to zoom into the figure. The
    ground truth trajectory is shown in cyan. Only in the zoomed excerpts, the
    map landmarks are visualized in white and a significant difference in
    density can be observed depending on the environment. An additional excerpt
    allows correlating the generated radar map with a third-party HD map. While
    a good general spatial agreement can be observed, numerous additional
    point-shaped landmarks are found in the radar map (mainly fence posts not
    included in HD map).}
  \label{fig:precfar_with_lm_map}
\end{figure*}

\noindent
The proposed localization framework is evaluated on three different routes in
Germany that are part of the cooperative project @CITY.
Table~\ref{tab:route_overview} gives an overview over the different routes.
Three passes per route are selected for evaluation. While the metric route
length refers to one pass, the evaluation duration includes all three passes.
Each route was driven in the same direction for every pass to eliminate any
possible influence on the study's main objective. The dataset is chosen to
include diverse scenarios, ranging from rural sections, over populated areas of
different density, to industrial parks. An example traversal of the Ulm route is
shown in Fig.~\ref{fig:precfar_with_lm_map} with an amplitude-only PreCFAR map
based on the ground truth trajectory and landmark positions of some sections.
Moreover, a comparison of radar map and HD map is shown.

\begin{figure*}
  \pgfkeys{/pgf/number format/.cd,1000 sep={}}
  \centering
  \input{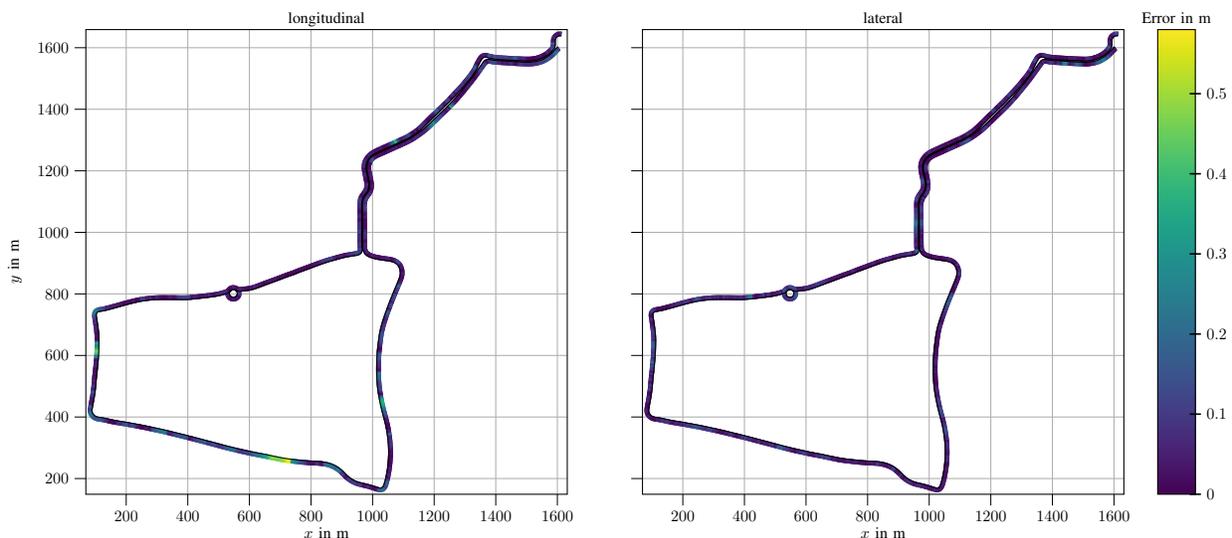}%
  \vspace{-0.15cm}
  \caption{The spatial distribution of the longitudinal and lateral trajectory
    errors of drive \#{}3 in Ulm allows identification of areas with increased
    errors. The generally better performance in the lateral compared to the
    longitudinal direction can also be observed in this example. By correlating
    the areas of low point-shaped landmark density depicted in
    Fig.~\ref{fig:precfar_with_lm_map}, the cause of larger localization errors
    can be understood.}
  \label{fig:lat_lon_error}
\end{figure*}

\subsection{Localization Accuracy}
\def\ate{\epsilon}
\def\atet{\mathbf{\ate}_{\textnormal{t}}}
\def\atetlat{\ate_\textnormal{lat}}
\def\atetlong{\ate_\textnormal{long}}
\def\ater{\ate_{\textnormal{rot}}}
\noindent
The quantitative evaluation of the localization result is based on absolute
trajectory errors. Assuming that post-processed GNSS/IMU data provides a ground
truth pose $\posenode_\textnormal{gt}$, the error to the estimated pose
$\hat{\posenode}$ is calculated. Let a pose $\posenode$ be expressed as a
pair of rotation $\mathbf{R} \in \liegrouprotation{2}$ and translation
$\mathbf{t} \in \mathbb{R}^2$. Then, the absolute trajectory errors of interest
are
\begin{itemize}
\item $\atet = \setlength\arraycolsep{2pt}\begin{pmatrix} \atetlong & \atetlat \end{pmatrix}^T =
  \mathbf{R}_\textnormal{gt}^{-1} \left( \hat{\mathbf{t}} -
    \mathbf{t}_\textnormal{gt} \right)$ for a translational and
\item $\ater = \arccos\left(0.5 \operatorname{tr} \left(
      \mathbf{R}_\textnormal{gt}^{-1}\hat{\mathbf{R}} \right) \right)$
  for a rotational part.
\end{itemize}
For better interpretation, the translational part is assessed by its components
in longitudinal and lateral directions $\atetlong$ and $\atetlat$, respectively.
The time series of trajectory errors is reduced to two scalar performance
figures for evaluation in the following: maximum and RMSE. In order to avoid any
influence from inaccuracies of the ground truth system, results are only
included in the performance figure calculation when a carrier phase solution is
available (cf.~Section~\ref{sec:measurement_setup}). Moreover, periods when the
vehicle comes to a standstill (more specifically $v \le
\SI{0.5}{\meter\per\second}$), e.g. at a traffic light, are not included to
avoid skewing the error distribution.

Table~\ref{tab:accuracy_table} gives an overview of the results for all drives
using fully polarimetric information. The column ``Reliable GNSS'' indicates how
often the precise ground truth solution was available. Comparing the
longitudinal and lateral results first, a difference can be observed. In
general, the lateral accuracy is better, especially in terms of RMSE. This can
be explained by the structure of typical automotive environments. The driving
lanes are typically bounded, e.g. by curbstones, guardrails or grass edges.
These boundaries are well detectable by radar, so the localization system can
orient itself to these objects to derive a good estimate of the lateral
position. As can be seen from the maximum lateral trajectory errors, it is
sufficient to infer the traveled lane in all evaluated drives.

\begin{table}
  \renewcommand{\arraystretch}{1.4}
  \caption{Localization accuracy using fully polarimetric information}
  \label{tab:accuracy_table}
  \centering
  \resizebox{\columnwidth}{!}{\begin{tabular}{lcrcccccc}
\toprule
\multirow{2}*{\bfseries Route} & \multirow{2}*{\bfseries Drive} & \multirow{2}*{\bfseries \makecell[c]{Reli.\\GNSS}} &\multicolumn{2}{c}{\bfseries longitudinal ($\atetlong$)} &\multicolumn{2}{c}{\bfseries lateral ($\atetlat$)} &\multicolumn{2}{c}{\bfseries rotational ($\ater$)} \\
& & & RMSE & max & RMSE & max & RMSE & max\\
\midrule
\multirow{3}*{Frankfurt}& \#1 & \SI{76}{\percent} & \SI{0.12}{\meter} & \SI{0.84}{\meter} & \SI{0.06}{\meter} & \SI{0.45}{\meter} & \SI{0.44}{\degree} & \SI{2.77}{\degree}\\
& \#2 & \SI{84}{\percent} & \SI{0.11}{\meter} & \SI{0.74}{\meter} & \SI{0.06}{\meter} & \SI{0.37}{\meter} & \SI{0.45}{\degree} & \SI{2.91}{\degree}\\
& \#3 & \SI{87}{\percent} & \SI{0.15}{\meter} & \SI{1.29}{\meter} & \SI{0.06}{\meter} & \SI{0.49}{\meter} & \SI{0.48}{\degree} & \SI{3.83}{\degree}\\
\midrule
\multirow{3}*{Ingolstadt}& \#1 & \SI{91}{\percent} & \SI{0.08}{\meter} & \SI{0.36}{\meter} & \SI{0.05}{\meter} & \SI{0.29}{\meter} & \SI{0.39}{\degree} & \SI{2.57}{\degree}\\
& \#2 & \SI{93}{\percent} & \SI{0.09}{\meter} & \SI{0.34}{\meter} & \SI{0.05}{\meter} & \SI{0.39}{\meter} & \SI{0.41}{\degree} & \SI{2.92}{\degree}\\
& \#3 & \SI{92}{\percent} & \SI{0.08}{\meter} & \SI{0.37}{\meter} & \SI{0.05}{\meter} & \SI{0.38}{\meter} & \SI{0.37}{\degree} & \SI{2.69}{\degree}\\
\midrule
\multirow{3}*{Ulm}& \#1 & \SI{87}{\percent} & \SI{0.13}{\meter} & \SI{0.76}{\meter} & \SI{0.06}{\meter} & \SI{0.46}{\meter} & \SI{0.46}{\degree} & \SI{2.67}{\degree}\\
& \#2 & \SI{85}{\percent} & \SI{0.13}{\meter} & \SI{0.73}{\meter} & \SI{0.06}{\meter} & \SI{0.25}{\meter} & \SI{0.45}{\degree} & \SI{1.81}{\degree}\\
& \#3 & \SI{85}{\percent} & \SI{0.10}{\meter} & \SI{0.43}{\meter} & \SI{0.06}{\meter} & \SI{0.34}{\meter} & \SI{0.44}{\degree} & \SI{2.30}{\degree}\\
\bottomrule
\end{tabular}
}%
\end{table}

Contrary to that, there might not always be enough infrastructure available to
allow a localization with comparable accuracy in longitudinal direction. An
example of a spatially resolved translational trajectory error in both
directions is shown in Fig.~\ref{fig:lat_lon_error}. In comparison to
Fig.~\ref{fig:precfar_with_lm_map}, a correlation between the areas of low
point-shaped landmark density and larger longitudinal trajectory errors can be
identified. While the lane boundaries allow for precise lateral localization,
the number of landmarks that help in longitudinal direction (point-shaped or
orthogonal line-shaped landmarks) is insufficient and the integration error of
the ego-motion starts to show. An improvement could be possible by refining the
radar-based ego-motion estimation or by adding other sensor modalities.
However, one can argue that this behavior only shows when the longitudinal
position is least critical. As soon as it becomes more critical for potential
turns, there is usually more infrastructure available for longitudinal
localization. %
Comparing the results of the routes in different cities among each other, a
significantly better result in terms of translational and rotational RMSEs as
well as the longitudinal maximum error can be observed for all drives in
Ingolstadt. The results' similarity of the three Ingolstadt drives leads to the
conclusion that this environment is the easiest one for localization. This is
due to the fact that a sufficient amount of localization-suitable infrastructure
is available everywhere along this route. The Frankfurt and Ulm routes, on the
other hand, have sections where this is not the case, resulting in worse
accuracies. As shown by the example in Fig.~\ref{fig:lat_lon_error}, this occurs
on connecting roads, where the low amount of infrastructure (guardrail posts or
guide posts) is occluded by low vegetation, so that a robust detection by radar
is not ensured.

\begin{figure}[b]
  \resizebox{\columnwidth}{!}{\input{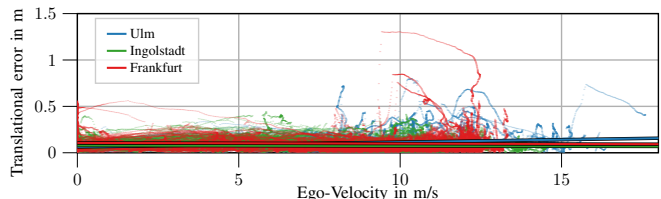}}\vspace{-0.2cm}
  \caption{The distribution of the translational localization error over all
    three drives per city shows an apparent positive correlation with the
    ego-velocity for the Ulm route, which is explainable by the landmark density
    in Fig.~\ref{fig:precfar_with_lm_map}.}
  \label{fig:speed_dependency}
\end{figure}

To assess the impact of the ego-velocity on the localization accuracy, the
translational error vector length is plotted over the ego-velocity in
Fig.~\ref{fig:speed_dependency}. The plotted linear regression does not show a
tendency of increased localization errors for higher velocities in Ingolstadt
and Frankfurt. In contrast, a small positive correlation is apparent for Ulm.
However, speed limits allow higher velocities only in the area of lower
landmark density as identified in Fig.~\ref{fig:precfar_with_lm_map}, such
that the apparent correlation is justified otherwise. Consequently, a general
degradation with increasing velocities can not be stated.

\subsection{Influence of Polarimetric Information}
\noindent
In order to compare the effect of different levels of polarization information
on the localization accuracy, the fully polarimetric measurement data was
sub-sampled in a number of experiments. While this omitting of some polarization
channels to simulate a radar with limited polarimetric information does not
preserve the back-scattered power (i.e. a non-polarimetric radar could have more
effective illumination time with the same overall measurement duration because
no TDM polarization switching is needed), the effect on the localization result
is expected to be small. This is assumed because low reflectivity landmark
candidates constitute no robust landmarks and thus offer negligible benefit for
localization. The much more decisive impact on exploiting lower reflectivity
landmarks is coming from leveraging PreCFAR gridmaps to avoid CFAR's masking
such that line-shaped candidates constitute a continuous area, which is robustly
identifiable by the extractor. An example of the negligible effect of the power
loss is given in Section~\ref{sec:line_lm_extraction} by the weak curbstone.

The comparison between the different polarization levels is done via
complementary cumulative distribution functions (CCDFs) of the rotational
trajectory error and of the translational errors in lateral and longitudinal
directions. It is defined as $\bar{F}\left(\epsilon\right) = 1 - F\left(
  \epsilon\right)$ %
with the cumulative distribution function $F\left( \epsilon\right)$.
$\bar{F}\left(\epsilon \right)$ is the probability that the trajectory error
will take a value greater than $\epsilon$. By using a logarithmic scaling of the
y-axis, it is easier to read the lower probability values for the larger
trajectory errors in the lower part of the plots.

Additional to the polarimetric information levels in the circular basis
$\lrbasis$, which could be derived by omitting elements, the channels of a
linear basis $\hvbasis$ were considered for evaluation. For this, the fully
polarimetric covariances in the circular basis were transformed to the linear
basis \cite{michelson_cov_circular} and subsampled from there if necessary.
Equation~\eqref{eq:wishart_distance} remains valid with the only change that the
covariances are now in the linear basis ($\polcovariancematrix_\hvbasis,
\neighborhoodcovariance_\hvbasis$).
\pagebreak

For the fully polarimetric case, one can show that the Wishart distance is the
same irrespective of which of the two bases is chosen. Therefore, the linear
fully polarimetric case is not shown in the following figures. Dual and single
polarimetric configurations in a linear basis are included in a way comparable
with the circular basis. Only the single cross-polarized case (HV or VH) is
dropped because of significantly worse results, which would have made
unfavorable rescaling of the figures necessary. This result is expected as
even-bounce scattering with the specific orientation required for a response in
these channels is rarely occurring at suitable landmarks.

Due to the covariance formulation in Eq.~\eqref{eq:covariance}, the absolute
phase is neglected in the single polarimetric experiments (LR or RR or HH) such
that only real amplitudes are used. Starting from two polarimetric
channels, the two-dimensional complex covariance is spanned and the relative
phases between channels are included additionally to the amplitudes by the
off-diagonal elements representing a part of the scattering characteristic.

\begin{figure}
  \resizebox{\columnwidth}{!}{\input{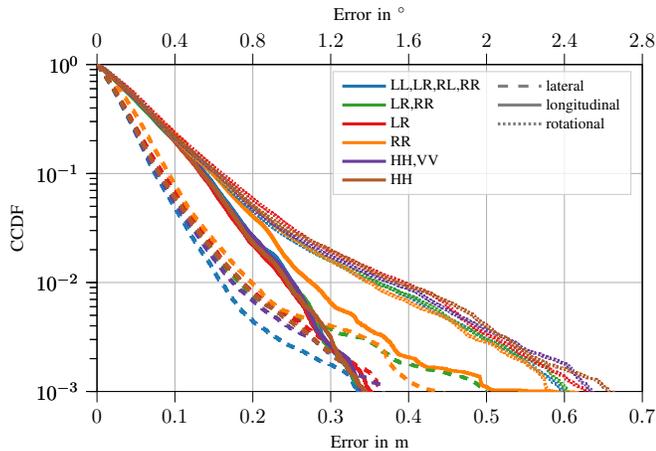}}%
  \caption{The distribution of trajectory errors accumulated over all three
    drives along the Ingolstadt route shows no significant performance difference
    depending on the polarization level used, except for the single
    polarization evaluation of the even-bounce channel RR.}
  \label{fig:ccdf_ingolstadt}
\end{figure}

In Fig.~\ref{fig:ccdf_ingolstadt}, the joint distribution over all three drives
along the Ingolstadt route is shown. Confirming the previous observation of
favorable results on this route, there is generally no significant degradation
of localization performance even with decreasing polarimetric information
content. This supports the hypothesis that a sufficient amount of
infrastructure is available and can already be detected by amplitude only.
Landmarks that require polarimetric information for reliable detection are not
needed due to the abundant presence of other landmarks. A minor exception is
observed for the even-bounce single polarization channel RR, which shows worse
errors in both translational dimensions. This is not surprising, because a large
part of the landmarks in automotive environments exhibit odd-bounce scattering
behavior, which makes a localization system relying only on even-bounce
scatterers the least preferred realization for such applications.

\begin{figure}[t]
  \resizebox{\columnwidth}{!}{\input{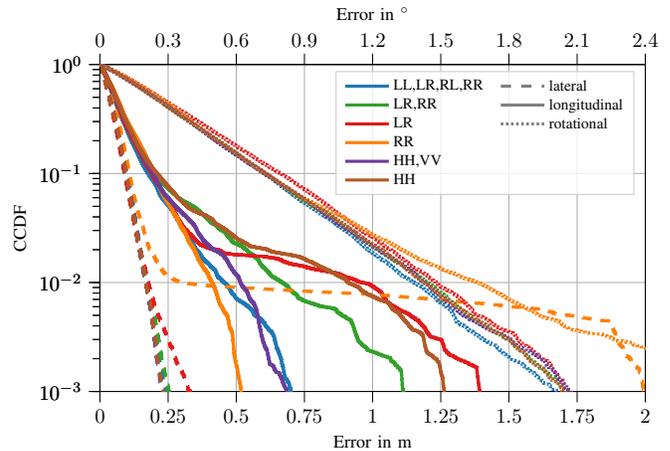}}%
  \caption{The distribution of trajectory errors accumulated over all three
    drives along the Ulm route shows a significant performance difference
    depending on the employed polarization level.}
  \label{fig:ccdf_ulm}
\end{figure}

Different results are found for the more challenging environment on the Ulm
routes, as is shown in Fig.~\ref{fig:ccdf_ulm}. Starting with the single
even-bounce channel RR, low errors are observed in longitudinal direction
compared to the other configurations on this route. This can be explained by the
fact that the low occurrence of landmarks with such scattering does not lead to
challenges with an unambiguous association. However, especially line-shaped
landmarks are rarely even-bounce scatterers that would have helped with the
lateral and rotational errors. Accordingly, the results are particularly poor in
these dimensions and this polarization configuration is the overall worst case.
For the remaining configurations, no remarkable difference in the lateral and
rotational errors is observed. A sufficient number of line-shaped landmarks
that are extracted by amplitude alone is the reason for this observation. This
supports the hypothesis that received power, which is most relevant for
extraction of continuous line-shaped landmarks, is not a limiting factor for
localization applications. Consequently, the approach of omitting channels can
be considered valid. Excluding RR, the longitudinal localization error decreases
with an increasing amount of polarimetric information. Contrary to the
Ingolstadt route, there is no longer an abundance of robust landmark extractions
irrespective of the polarimetric configuration used. The addition of scattering
uniqueness as a criterion during landmark extraction seems to offer the
expected advantage in this environment. Further evidence that the advantage is
not just caused by the ability to receive power from even- and odd-bounce
objects can be derived by considering the linear basis results. With the HH
configuration, even- and odd-bounce scattering of most relevant objects
(excluding rarely occurring \SI{45}{\degree} rotated dihedral structures) is
received but the performance is still comparable to the inferior LR result. Only
the linear dual polarimetric configuration HH,VV shows a significant
improvement. Although no reception of other polarizations is added, the ability
of a scattering-based extraction by distinguishing between even- and odd-bounce
scattering instead of a purely amplitude-based approach enables the improvement.

\section{CONCLUSION}
\label{sec:conclusion}
\noindent
This paper presented a radar-only landmark-based localization approach that
exploits polarimetric scattering information for a more robust detection of
landmarks. Extensive real-world experiments with more than \SI{2.5}{\hour} of
driving demonstrated the high accuracy of the localization in diverse
environments.
When using fully polarimetric data, average RMS errors of
\SI{0.11}{\meter}, \SI{0.06}{\meter} and \SI{0.43}{\degree} are observed for
longitudinal, lateral and rotational components, respectively. It was also shown
how the localization performance benefits from adding increasing amounts of
polarimetric information in challenging scenarios. A limitation of the approach
is that the map construction relies on the availability of a ground truth pose.
Therefore, future work involves constructing the maps at least partially in a
SLAM fashion when no ground truth is available, and studying how this affects
the localization performance.

\bibliographystyle{IEEEtran}
\bibliography{references}
\vspace{-0.1cm}
\begin{IEEEbiography}[{\includegraphics[width=1in, height=1.25in, clip, keepaspectratio]{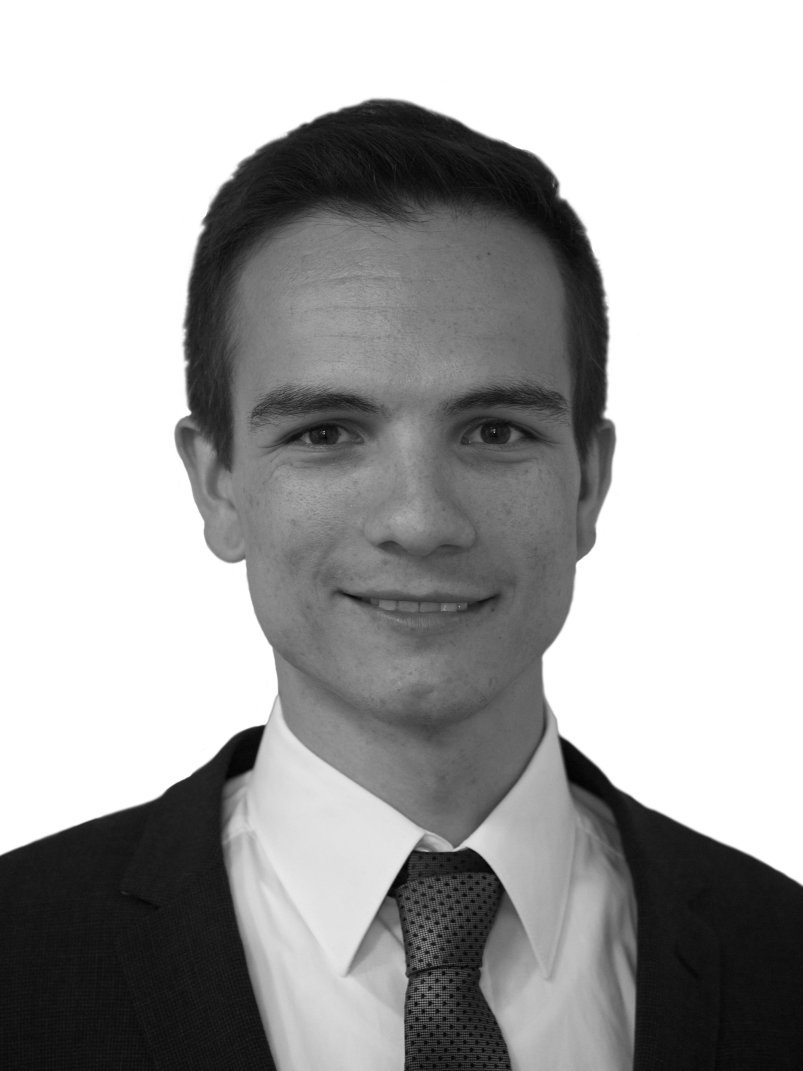}}]
{Fabio Weishaupt} received his B.Sc. and M.Sc. degrees in electrical engineering
from RWTH Aachen University, Aachen, Germany, in 2015 and 2017, respectively.
His master thesis was carried out at Fraunhofer FHR, Wachtberg, Germany, on
vital sign measurements with mmWave radar. Since 2018, he is working towards a
Ph.D. degree on the topic of polarimetric radar for automotive self-localization
at the research and development department of Mercedes-Benz AG together with
RWTH Aachen University, Aachen, Germany. His research interests include mmWave
system design, polarimetry, radar signal processing and perception algorithms.
\end{IEEEbiography}%

\begin{IEEEbiography}[{\includegraphics[width=1in, height=1.25in, clip, keepaspectratio]{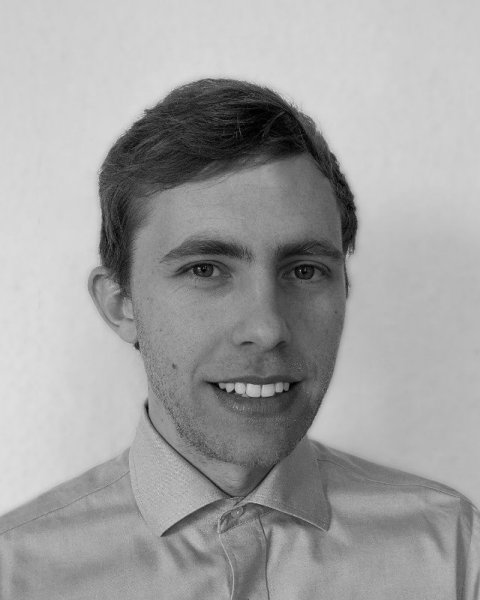}}]
  {Julius F. Tilly} received his master’s degree in Physics from Technical
  University of Dortmund in 2017. In his master’s thesis he used Nuclear
  Magnetic Resonance spectroscopy to investigate ion dynamics in lithium borate
  glass and sodium ion conductors.
  Since 2018, he is a PhD student at Mercedes-Benz AG in the research and
  development department for environment perception. He is researching in the
  field of road user classification and tracking with radars for autonomous
  driving. His research interests include polarimetric radars, radar signal
  processing and algorithms for environment perception with radars.
\end{IEEEbiography}%
\vskip -1.5\baselineskip plus -1fil
\begin{IEEEbiography}[{\includegraphics[width=1in, height=1.25in, clip, keepaspectratio]{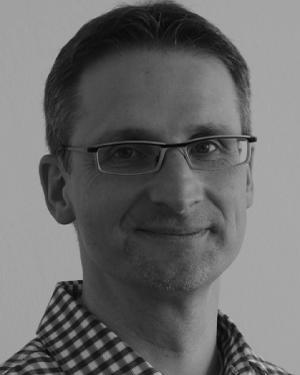}}]
  {Nils Appenrodt} received his Dipl.-Ing. degree in electrical engineering from
  University Duisburg, Germany, in 1996. He was a research assistant in the
  field of imaging radar systems with University Duisburg. Since 2000, he has
  been with the group research and advanced engineering, formerly Daimler AG now
  Mercedes-Benz AG, as a research engineer and manager, where he is working in
  the field of environment perception systems. His research interests include
  radar sensor processing, sensor data fusion, advanced driver assistance and autonomous
  driving systems.
\end{IEEEbiography}%
\vskip -1.5\baselineskip plus -1fil
\vspace{0.1cm}
\begin{IEEEbiography}[{\includegraphics[width=1in, height=1.25in, clip, keepaspectratio]{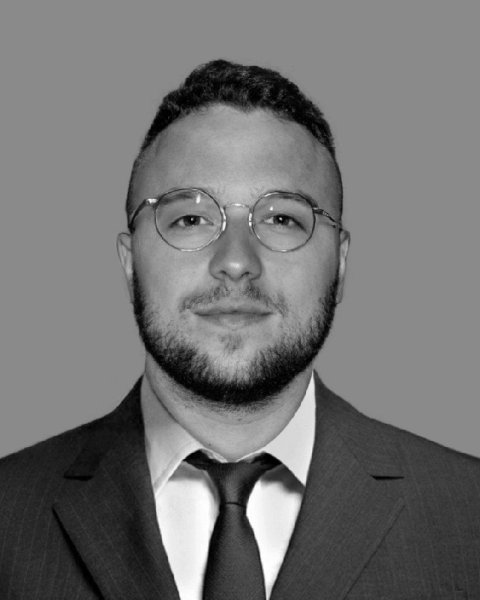}}]
  {Pascal Fischer} received his B.Sc. and M.Sc. degrees in computer science with
  specialization in artificial intelligence from Goethe University, Frankfurt am
  Main, Germany in 2020 and 2021, respectively. His master thesis was carried
  out at Mercedes-Benz AG, Stuttgart, Germany, on automotive self-localization
  with polarimetric radar. Since the end of 2021, he is working as a machine
  learning engineering consultant at Accenture in various industries.
\end{IEEEbiography}%
\vskip -1.5\baselineskip plus -1fil
\vspace{0.1cm}
\begin{IEEEbiography}[{\includegraphics[width=1in, height=1.25in, clip, keepaspectratio]{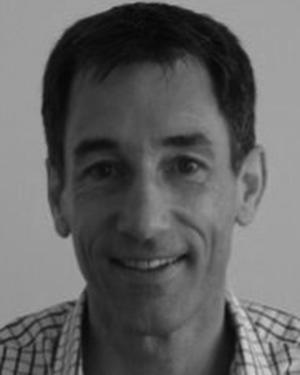}}]
  {J\"urgen Dickmann} received the Diploma degree in electrical engineering from
  University Duisburg, Duisburg, Germany, in 1984.  He received the Dr.-Ing.
  degree from Rheinisch Westfaelische Technische Hochschule Aachen (RWTH),
  Aachen, Germany.  He is the Head of radar sensors and radar-based perception
  at autonomous driving, Mercedes-Benz AG. In this role, he is responsible for
  research to serial development of radar and radar-based environmental
  understanding for autonomous driving for all platforms inside Mercedes-Benz
  AG. Among others, he held manager positions at Daimler AG in laser-scanner,
  sensor-fusion, and situation analysis. He conducted radar developments for
  radar-based precrash and driver assistant systems for all platforms (passenger
  cars, busses, van, truck) inside Daimler AG, including E/S-Class.
  In 1986, he started his career at AEG Research Center, where he did
  research on III/V-semiconductor processing techniques, mm-Wave device, and
  MMIC-design (up to 120 GHz) and fabrication.
\end{IEEEbiography}%
\vskip -1.5\baselineskip plus -1fil
\vspace{0.1cm}
\begin{IEEEbiography}[{\includegraphics[width=1in, height=1.25in, clip, keepaspectratio]{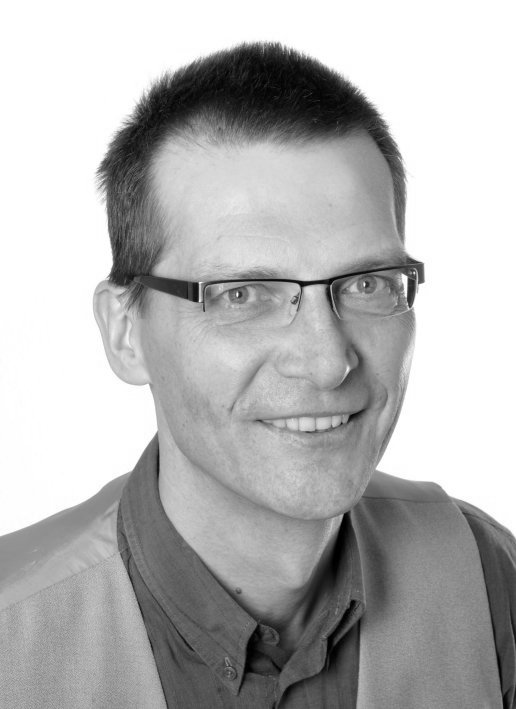}}]
{Dirk Heberling} [M'03, SM'10] received his doctoral degree (Dr.-Ing.) in 1993.
In 1993, he joined IMST GmbH, Kamp-Lintfort, Germany, to establish a new antenna
section, and from 1995 to 2003, he was the head of the Antennas Department. He
has been a member of the European Competence Projects for Antennas COST 260,
COST 284, IC0603, and IC1102. From 2003 to 2008, he took over the Department of
Information and Communication Systems of IMST GmbH, and in 2008, he moved to
RWTH Aachen, where he is the head of the institute and chair for High Frequency
Technology. He is a member of VDE and is the German delegate to IC1102. He is a
member of the Steering Committee and Organizing Committee for the European
Conference on Antennas and Propagation. In 2016, he became the head of the
Fraunhofer Institute for High Frequency Physics and Radar Techniques FHR.
\end{IEEEbiography}

\end{document}